%% file: main.tex
\documentclass[12pt]{article}
\usepackage[utf8]{inputenc}

\usepackage{pdfsync}
\usepackage{fullpage}
\newif\ifdraft \drafttrue
\newif\iffull \fulltrue

\usepackage{graphicx}
\usepackage{xcolor}
\usepackage[english]{babel}
\usepackage{csquotes}

\usepackage{xspace}
\usepackage[T1]{fontenc}

\usepackage[numbers]{natbib}

\usepackage[utf8]{inputenc} 
\usepackage[T1]{fontenc}    
\usepackage{hyperref}       
\usepackage{url}            
\usepackage{booktabs}       
\usepackage{amsfonts}       
\usepackage{nicefrac}       
\usepackage{microtype}      
\usepackage{xcolor}         

\usepackage[ruled,vlined]{algorithm2e}
\usepackage{amsmath}
\usepackage{amssymb}
\usepackage{amsthm}
\usepackage{graphicx}
\usepackage{subcaption}
\usepackage{tabularx}
\usepackage{multirow}
\usepackage{multicol}
\usepackage{apxproof}
\usepackage{cleveref}
\usepackage{color-edits}
\usepackage{bbm}

\addauthor{sw}{blue}

\input{math_helpers}

\newtheoremrep{inequality}{Inequality}
\newtheoremrep{theorem}{Theorem}
\newtheoremrep{lemma}{Lemma}
\newtheoremrep{definition}{Definition}
\newtheoremrep{proposition}{Proposition}

\newcommand{\hdmm}{\mbox{{\sf HDMM}}\xspace}
\newcommand{\hdmmpgm}{\mbox{{\sf HDMM+PGM}}\xspace}
\newcommand{\dq}{\mbox{{\sf DualQuery}}\xspace}
\newcommand{\fem}{\mbox{\sf FEM}\xspace}
\newcommand{\mwem}{\mbox{{\sf MWEM}}\xspace}
\newcommand{\pmwpub}{\mbox{{\sf PMW\textsuperscript{Pub}}}\xspace}

\newcommand{\expmech}{\mbox{EM}\xspace}
\newcommand{\gaussian}{\mbox{G}\xspace}

\newcommand{\adaptive}{\mbox{{\sf Adaptive Measurements}}\xspace}
\newcommand{\rap}{\mbox{{\sf RAP}}\xspace}
\newcommand{\rapsoftmax}{\mbox{{\sf RAP\textsuperscript{softmax}}}\xspace}

\newcommand{\pep}{\mbox{{\sf PEP}}\xspace}
\newcommand{\peppub}{\mbox{{\sf PEP\textsuperscript{Pub}}}\xspace}
\newcommand{\gem}{\mbox{{\sf GEM}}\xspace}
\newcommand{\gemupdate}{\mbox{{\sf GEM-UPDATE}}\xspace}
\newcommand{\gempub}{\mbox{{\sf GEM\textsuperscript{Pub}}}\xspace}
\newcommand{\gempubred}{\mbox{{\sf GEM\textsuperscript{Pub} (reduced)}}\xspace}

\newcommand{\ema}{\mbox{{\sf EMA}\xspace}}

\newcommand{\tl}[1]{ {\color{blue}TL: #1}}
\newcommand{\gv}[1]{ {\color{red}GV: #1}}

\allowdisplaybreaks

\begin{document}

\title{Iterative Methods for Private Synthetic Data:\\ Unifying Framework and New Methods}

\author{
Terrance Liu
\thanks{First two authors contributed equally.} 
\thanks{Carnegie Mellon University ~\dotfill~ \texttt{terrancl@andrew.cmu.edu}}
\\
\and 
Giuseppe Vietri
\footnotemark[1]
\thanks{University of Minnesota ~\dotfill~ \texttt{vietr002@umn.edu}}
\\
\and
Zhiwei Steven Wu
\thanks{Carnegie Mellon University ~\dotfill~ \texttt{zstevenwu@cmu.edu}}
}

\date{}

\maketitle


\begin{abstract}

We study private synthetic data generation for query release, where the goal is to construct a sanitized version of a sensitive dataset, subject to differential privacy, that approximately preserves the answers to a large collection of statistical queries. We first present an algorithmic framework that unifies a long line of iterative algorithms in the literature. Under this framework, we propose two new methods. The first method, private entropy projection (\pep), can be viewed as an advanced variant of \mwem that adaptively reuses past query measurements to boost accuracy. Our second method, generative networks with the exponential mechanism (\gem), circumvents computational bottlenecks in algorithms such as \mwem and \pep by optimizing over generative models parameterized by neural networks, which capture a rich family of distributions while enabling fast gradient-based optimization. We demonstrate that \pep and \gem empirically outperform existing algorithms. Furthermore, we show that \gem nicely incorporates prior information from public data while overcoming limitations of \pmwpub, the existing state-of-the-art method that also leverages public data.

\end{abstract}


\input{docs/intro}

\input{docs/prelims}

\input{docs/framework}

\input{docs/methods/pep}
\input{docs/methods/gem}
\input{docs/methods/methods_public}

\input{docs/experiments/experiments}

\section{Conclusion}

In this work, we present a framework that unifies a long line of iterative private query release algorithms by reducing each method to a choice of some distributional family $\cD$ and loss function $\cL$. We then develop two new algorithms, \pep and \gem, that outperform existing query release algorithms. In particular, we empirically validate that \gem performs very strongly in high dimensional settings (both with and without public data). We note that we chose a rather simple neural network architecture for \gem, and so for future work, we hope to develop architectures more tailored to our problem. Furthermore, we hope to extend our algorithms to other query classes, including mixed query classes and convex minimization problems \citep{jonconvex}.



\section*{Acknowledgments} 
ZSW is supported by NSF grant SCC-1952085, Carnegie Mellon CyLab’s Secure and Private IoT Initiative, and a Google Faculty Research Award.

\bibliography{main}
\bibliographystyle{plainnat}

\newpage
\input{docs/appendix/appendix}

\end{document}

%% file: math_helpers.tex
\newtheorem{theorem}{Theorem}
\newtheorem{lemma}{Lemma}

\DeclareMathOperator{\argmax}{\text{arg}\max}
\DeclareMathOperator{\argmin}{\text{arg}\min}
\newcommand{\cX}{\mathcal{X}}
\newcommand{\pr}[1]{\mathrm{Pr}\left[#1 \right]}
\newcommand{\diver}[1]{\text{RE}\left(#1 \right)}
\newcommand{\rendiver}[2]{\mathrm{R}_{#1}\left(#2 \right)}

\newcommand{\eps}{\varepsilon}
\newcommand{\cM}{\mathcal{M}}
\newcommand{\cN}{\mathcal{N}}

\newcommand{\cL}{\mathcal{L}}
\newcommand{\cQ}{\mathcal{Q}}

\newcommand{\cD}{\mathcal{D}}

\newcommand{\cR}{\mathcal{R}}

\newcommand{\atil}{\widetilde{a}}
\newcommand{\Atil}{\widetilde{A}}
\newcommand{\qtil}{\widetilde{q}}
\newcommand{\Qtil}{\widetilde{Q}}

\newcommand{\ahat}{\widehat{a}}

\newcommand{\Dpriv}{P}

\newcommand{\abs}[1]{\left| #1 \right|}

\newcommand{\pp}[1]{\left(#1\right)}

\newcommand{\roundbrack}[1]{\left( #1 \right)}
\newcommand{\curlybrack}[1]{\left\lbrace #1 \right\rbrace}
\newcommand{\squarebrack}[1]{\left\lbrack #1 \right\rbrack}
\newcommand{\anglebrack}[1]{\left\langle #1 \right\rangle}

\newcommand{\vect}[1]{\mathbf{#1}}

%% file: docs/intro.tex
\section{Introduction}

As the collection and analyses of sensitive data become more prevalent, there is an increasing need to protect individuals' private information. Differential privacy \cite{DworkMNS06} is a rigorous and meaningful criterion for privacy preservation that enables quantifiable trade-offs between privacy and accuracy. In recent years, there has been a wave of practical deployments of differential privacy across organizations such as Google, Apple, and most notably, the U.S. Census Bureau \cite{Abowd18}.

In this paper, we study the problem of differentially private query release: given a large collection of statistical queries, the goal is to release approximate answers subject to the constraint of differential privacy. Query release has been one of the most fundamental and practically relevant problems in differential privacy. For example, the release of summary data from the 2020 U.S. Decennial Census can be framed as a query release problem. We focus on the approach of synthetic data generation---that is, generate a privacy-preserving "fake" dataset, or more generally a representation of a probability distribution, that approximates all statistical queries of interest. Compared to simple Gaussian or Laplace mechanisms that perturb the answers directly, synthetic data methods can provably answer an exponentially larger collection of queries with non-trivial accuracy. However, their statistical advantage also comes with a computational cost. Prior work has shown that achieving better accuracy than simple Gaussian perturbation is intractable in the worst case even for the simple query class of 2-way marginals that release the marginal distributions for all pairs of attributes \citep{UllmanV11}.

Despite its worst-case intractability, there has been a recent surge of work on practical algorithms for generating private synthetic data. Even though they differ substantially in details, these algorithms share the same iterative form that maintains and improves a probability distribution over the data domain: identifying a small collection of high-error queries each round and updating the distribution to reduce these errors. Inspired by this observation, we present a unifying algorithmic framework that captures these methods. Furthermore, we develop two new algorithms, \gem and \pep, and extend the former to the setting in which public data is available. We summarize our contributions below:

\paragraph{Unifying algorithmic framework.}
We provide a framework that captures existing iterative algorithms and their variations. At a high level, algorithms under this framework maintain a probability distribution over the data domain and improve it over rounds by optimizing a given loss function. We therefore argue that under this framework, the optimization procedures of each method can be reduced to what loss function is minimized and how its distributional family is parameterized. For example, we can recover existing methods by specifying choices of loss functions---we rederive \mwem \citep{hardt2010multiplicative} using an entropy-regularized linear loss, \fem \cite{vietri2020new} using a linear loss with a linear perturbation, and \dq \citep{gaboardi2014dual} with a simple linear loss. Lastly, our framework lends itself naturally to a softmax variant of \rap \citep{aydore2021differentially}, which we show outperforms \rap itself.\footnote{We note that \citet{aydore2021differentially} have since updated the original version (\url{https://arxiv.org/pdf/2103.06641v1.pdf}) of their work to include a modified version of \rap that leverages SparseMax \citep{martins2016softmax}, similar to way in which the softmax function is applied in our proposed baseline, \rapsoftmax.}

\paragraph{Generative networks with the exponential mechanism (\gem).} \gem is inspired by \mwem, which attains worst-case theoretical guarantees that are nearly information-theoretically optimal \citep{bun2018fingerprinting}. However, \mwem maintains a joint distribution over the data domain, resulting in a runtime that is exponential in the dimension of the data. \gem avoids this fundamental issue by optimizing the absolute loss over a set of generative models parameterized by neural networks. We empirically demonstrate that in the high-dimensional regime, \gem outperforms all competing methods.

\paragraph{Private Entropy Projection (\pep).} The second algorithm we propose is \pep, which can be viewed as a more advanced version of \mwem with an adaptive and optimized learning rate. We show that \pep minimizes a regularized exponential loss function that can be efficiently optimized using an iterative procedure. Moreover, we show that \pep monotonically decreases the error over rounds and empirically find that it achieves higher accuracy and faster convergence than \mwem.

\paragraph{Incorporating public data.} 
Finally, we consider extensions of our methods that incorporate prior information in publicly available datasets (e.g., previous data releases from the American Community Survey (ACS) prior to their differential privacy deployment). While \citet{liu2021leveraging} has established \pmwpub as a state-of-the-art method for incorporating public data into private query release, we discuss how limitations of their algorithm prevent \pmwpub from effectively using certain public datasets. We then demonstrate empirically that \gem circumvents such issues via simple pretraining, achieving max errors on 2018 ACS data for Pennsylvania (at $\varepsilon=1$) \textbf{9.23x} lower than \pmwpub when using 2018 ACS data for California as the public dataset.


\subsection{Related work}
Beginning with the seminal work of \citet{BlumLR08}, a long line of theoretical work has studied private synthetic data for query release \citep{RothR10, hardt2010multiplicative, hardt2010simple, GRU}. While this body of work establishes optimal statistical rates for this problem, their proposed algorithms, including \mwem \cite{hardt2010simple}, typically have running time exponential in the dimension of the data. While the worst-case exponential running time is necessary (given known lower bounds \citep{complexityofsyn, ullman13, ullmanpcp}), a recent line of work on practical algorithms leverage optimization heuristics to tackle such computational bottlenecks \cite{gaboardi2014dual, vietri2020new, aydore2021differentially}. In particular, \dq \cite{gaboardi2014dual} and \fem \cite{vietri2020new} leverage integer program solvers to solve their NP-hard subroutines, and \rap \cite{aydore2021differentially} uses gradient-based methods to solve its projection step. In Section \ref{sec:adaptive}, we demonstrate how these algorithms can be viewed as special cases of our algorithmic framework. Our work also relates to a growing line of work that use public data for private data analyses 
\cite{bassily2020private, alonlimits, BassilyMN20}. For query release, our algorithm, \gem, improves upon the state-of-the-art method, \pmwpub \citep{liu2021leveraging}, which is more limited in the range of public datasets it can utilize. Finally, our method \gem is related to a line of work on differentially private GANs \cite{beaulieu2019privacy, yoon2018pategan, neunhoeffer2020private, rmsprop_DPGAN}. However, these methods focus on generating synthetic data for simple downstream machine learning tasks rather than for query release.

Beyond synthetic data, a line of work on query release studies "data-independent" mechanisms (a term formulated in \citet{ENU20}) that perturb the query answers with noise drawn from a data-independent distribution (that may depend on the query class). This class of algorithms includes the matrix mechanism \citep{mm}, the high-dimensional matrix mechanism (\hdmm) \citep{mckenna2018optimizing}, the projection mechanism \citep{NTZ}, and more generally the class of factorization mechanisms \citep{ENU20}. In addition, \citet{mckenna2019graphical} provide an algorithm that can further reduce query error by learning a probabilistic graphical model based on the noisy query answers released by privacy mechanisms.

%% file: docs/prelims.tex
\section{Preliminaries}
Let $\cX$ denote a finite $d$-dimensional data domain (e.g., $\cX = \{0, 1\}^d$). 
Lete $U$ be the uniform distribution over the domain $\cX$. Throughout this work, we assume a private dataset $P$ that contains the data of $n$ individuals. For any $x\in\cX$, we represent $P(x)$ as the normalized frequency of $x$ in dataset $P$ such that $\sum_{x\in \cX} P(x) = 1$. One can think of a dataset $P$ either as a multi-set of items from $\cX$ or as a distribution over $\cX$.

We consider the problem of accurately answering an extensive collection of linear statistical queries (also known as counting queries) about a dataset. Given a finite set of queries $Q$, our goal is to find a synthetic dataset $D$ such that the maximum error over all queries in $Q$, defined as $\max_{q\in Q} | q(P)-q(D) |$, is as small as possible. For example, one may query a dataset by asking the following: how many people in a dataset have brown eyes? More formally, a statistical linear query $q_\phi$  is defined by a predicate function $\phi:\cX \rightarrow \{0,1\}$, as $q_\phi(D) = \sum_{x\in\cX}\phi(x)D(x)$ for any normalized dataset $D$. Below, we define an important, and general class of linear statistical queries called $k$-way marginals.

\begin{definition}[$k$-way marginal] \label{def:marginals}
%
Let the data universe with $d$ categorical attributes be $\mathcal{X}=\left(\mathcal{X}_{1} \times \ldots \times \mathcal{X}_{d}\right)$, where each $\mathcal{X}_{i}$ is the discrete domain of the $i$th attribute $A_i$. 
A $k$-way marginal query is defined by a subset $S\subseteq [d] $ of $k$ features (i.e., $|S| = k$) plus a target value $y\in \prod_{i\in S} \cX_i$ for each feature in $S$. Then the marginal query $\phi_{S,y}(x)$ is given by:
\begin{align*}
    \phi_{S,y}(x)= 
    \prod_{i\in S} \mathbbm{1}\pp{x_i = y_i}
\end{align*}
where $x_{i} \in \mathcal{X}_{i}$ means the $i$-th attribute of record $x \in \mathcal{X}$. Each marginal has a total of $\prod_{i=1}^{k} \left|\mathcal{X}_{i}\right|$ queries, and we define a \textit{workload} as a set of marginal queries. 

\end{definition}

We consider algorithms that input a dataset $P$ and produce randomized outputs that depend on the data.  The output of a randomized mechanism $\cM:\cX^*\rightarrow \cR$ is a privacy preserving computation if it satisfies differential privacy (DP) \cite{DworkMNS06}. We say that two datasets are neighboring if they differ in at most the data of one individual.

\begin{definition}[Differential privacy \cite{DworkMNS06}]
A randomized mechanism $\cM:\cX^n \rightarrow \cR$ is $(\eps,\delta)$-differentially privacy, if for all neighboring datasets $P, P'$ (i.e., differing on a single person), and all measurable subsets $S\subseteq \cR$ we have:
\begin{align*}
    \pr{ \cM(P) \in S} \leq e^{\eps} \pr{\cM(P') \in S} + \delta
\end{align*}
\end{definition}

Finally, a related notion of privacy is called concentrated differential privacy (zCDP) \cite{DworkR16,BunS16}, which enables cleaner composition analyses for privacy. 
%
%
\begin{definition}[Concentrated DP,  \citet{DworkR16,BunS16}]\label{def:zCDP}
 A randomized mechanism $\cM:\cX^n \rightarrow \cR$ is $\frac12\tilde{\eps}^2$-CDP, if for all neighboring datasets $P, P'$ (i.e., differing on a single person), and for all $\alpha\in(1,\infty)$,
 \begin{align*}
     \rendiver{\alpha}{\cM(P)\parallel\cM(P')} \leq \frac 12\tilde{\eps}^2\alpha
 \end{align*}
 where $\rendiver{\alpha}{\cM(P)\parallel\cM(P')}$ is the R\'{e}nyi divergence between the distributions $\cM(P)$ and $\cM(P')$.
\end{definition}

%% file: docs/framework.tex
\section{A Unifying Framework for Private Query Release}
\label{sec:adaptive}

%
%

In this work, we consider the problem of finding a distribution in some family of distributions $\cD$ that achieves low error on all queries. More formally, given a private dataset $P$ and a query set $Q$, we solve an optimization problem of the form:
\begin{align}\label{eq:generalproblem}
    \min_{D\in \cD} \max_{q\in Q} | q(P) - q(D)|
\end{align}

In Algorithm \ref{alg:adaptive}, we introduce \adaptive, which serves as a general framework for solving \eqref{eq:generalproblem}. At each round $t$, the framework uses a private selection mechanism to choose $k$ queries $\Qtil_t = \{\qtil_{t,1}, \ldots, \qtil_{t,k} \}$ with higher error from the set $Q$. It then obtains noisy measurements for the queries, which we denote by $\Atil_t =\{\atil_{t,1},\ldots, \atil_{t,k}\}$, where $\atil_{t,i} = \qtil_{t,i} + z_{t,i}$ and $z_{t,i}$ is random Laplace or Gaussian noise. Finally, it updates its approximating distribution $D_t$, subject to a loss function $\cL$ that depends on $\Qtil_{1:t} =\bigcup_{i=1}^t \Qtil_i$ and $\Atil_{1:t}=\bigcup_{i=1}^t\Atil_i$. We note that $\cL$ serves as a surrogate problem to \eqref{eq:generalproblem} in the sense that the solution of $\min_{D\in\cD} \cL(D)$ is an approximate solution for \eqref{eq:generalproblem}. We list below the corresponding loss functions for various algorithms in the literature of differentially private synthetic data. (We defer the derivation of these loss functions to the appendix.)

\paragraph{\mwem from \citet{hardt2010simple}} 
\mwem solves an entropy regularized minimization problem: 
\begin{align*}
    \cL^{\mwem}(D, \Qtil_{1:t}, \Atil_{1:t})
    = \sum_{i=1}^t \sum_{x\in \cX} D(x) \qtil_i(x) \pp{\atil_i - \qtil_i(D_{i-1})}
    + \sum_{x\in\cX} D(x) \log D(x)
\end{align*}
We note that \pmwpub \citep{liu2021leveraging} optimizes the same problem but restricts $\cD$ to distributions over the public data domain while initializing $D_0$ to be the public data distribution.

\paragraph{\dq from \citet{gaboardi2014dual}}
At each round $t$, \dq samples $s$ queries ($\Qtil_t = \{\qtil_{t,1}, \ldots \qtil_{t,s}\}$) from $\cQ_t$ and outputs $D_t$ that minimizes the the following loss function:
\begin{align*}
    \cL^{\dq}(D, \Qtil_t)
    = \sum_{i=1}^{s}  \qtil_{t,i}(D)
\end{align*}

\paragraph{\fem from \citet{vietri2020new}}
The algorithm \fem employs a follow the perturbed leader strategy, where on round $t$, \fem chooses the next distribution by solving:
\begin{align*}
  \cL^{\fem}(D, \Qtil_{1:t}) = 
    \sum_{i=1}^t \qtil_t(D) + \mathbb{E}_{x\sim D, \eta\sim \text{Exp}(\sigma)^d}\pp{\langle x, \eta \rangle} 
\end{align*}

\paragraph{\rapsoftmax adapted from \citet{aydore2021differentially}}
We note that \rap follows the basic structure of \adaptive, where at iteration $t$, \rap solves the following optimization problem:
\begin{align*}
    \cL^{\rap}(D, \Qtil_{1:t}, \Atil_{1:t})
    = \sum_{i, j} \roundbrack{\qtil_{i,j}(D) - \atil_{i,j}}^2
\end{align*}
However, rather than outputting a dataset that can be expressed as some distribution over $\cX$, \rap projects the noisy measurements onto a continuous relaxation of the binarized feature space of $\cX$, outputting $D \in [-1, 1]^{n' \times d}$ (where $n'$ is an additional parameter). Therefore to adapt \rap to \adaptive, we propose a new baseline algorithm that applies the softmax function instead of clipping each dimension of $D$ to be between $-1$ and $1$. For more details, refer to Section \ref{sec:gem} and Appendix \ref{appx:gem}, where describe how softmax is applied in \gem in the same way. With this slight modification, this algorithm, which we denote as \rapsoftmax, fits nicely into the \adaptive framework in which we output a synthetic dataset drawn from some probabilistic family of distributions $\cD = \curlybrack{\sigma(M) | M \in \mathbb{R}^{n' \times d}}$.

\input{docs/algos/adaptive}

Finally, we note that in addition to the loss function $\cL$, a key component that differentiates algorithms under this framework is the distributional family $\cD$ that the output of each algorithm belongs to. We refer readers to Appendix \ref{appx:loss_functions}, where we describe in more detail how existing algorithms fit into our general framework under different choices of $\cL$ and $\cD$.

\input{docs/privacy}

%% file: docs/algos/adaptive.tex
\begin{algorithm}[H]
\SetAlgoLined

\textbf{Input:} Private dataset $\Dpriv$ with $n$ records, set of linear queries $Q$, distributional family $\cD$,  loss functions $\cL$, number of iterations $T$ \\
Initialize distribution $D_0 \in \cD$ \\
 \For{$t = 1, \ldots, T$}{
  \textbf{Sample}:
  For $i\in[k]$, choose $\qtil_{t,i}$ using a differentially private selection mechanism. \\
  \textbf{Measure:} 
  For $i\in[k]$, let $\atil_{t,i} = \qtil_{t,i}(\Dpriv) + z_{t,i} $ where $z$ is Gaussian or Laplace noise \\
  \textbf{Update:}
  Let $\Qtil_t = \{\qtil_{t,1}, \ldots, \qtil_{t,k}\}$ and $\Atil_t =\{\atil_{t,1},\ldots, \atil_{t,k}\}$. Update distribution $D$:
  \begin{align*}
    D_{t} \leftarrow \argmin_{D\in \cD} \cL\pp{D_{t-1}, \Qtil_{1:t}, \Atil_{1:t}}
  \end{align*}
  where
  $\Qtil_{1:t} = \bigcup_{i=1}^t \Qtil_{i}$ and $\Atil_{1:t} = \bigcup_{i=1}^t \Atil_i$.
  
 }
 Output $H\roundbrack{\curlybrack{D_t}_{t=0}^{T}}$  where $H$ is some function over all distributions $D_t$ (such as the average)
 \caption{\adaptive}
 \label{alg:adaptive}
\end{algorithm}

%% file: docs/privacy.tex
\subsection{Privacy analysis}\label{sec:privacy}

We present the privacy analysis of the \adaptive framework while assuming that the exponential and Gaussian mechanism are used for the private \textit{sample} and noisy \textit{measure} steps respectively. More specifically, 
suppose that we (1) sample $k$ queries using the \textit{exponential mechanism} with the score function:
\begin{align*}
    \pr{\qtil_{t,i} = q} \propto \exp\roundbrack{\alpha \varepsilon_0 n |q(\Dpriv) - q(D_{t-1})|}
\end{align*}
and (2) measure the answer to each query by adding Gaussian noise 
\begin{align*}
    z_{t,i} \sim \mathcal{N} \roundbrack{0, \roundbrack{\frac{1}{n(1-\alpha)\varepsilon_0}}^2}.
\end{align*}
Letting $\varepsilon_0 = \sqrt{\frac{2\rho}{T \roundbrack{\alpha^2 + \roundbrack{1-\alpha}^2}}}$ and $\alpha \in (0, 1)$ be a privacy allocation hyperparameter (higher values of $\alpha$ allocate more privacy budget to the exponential mechanism), we present the following theorem:

\begin{theorem}
When run with privacy parameter $\rho$, \adaptive satisfies $\rho$-zCDP. Moreover for all $\delta>0$, \adaptive satisfies$\left( \varepsilon(\delta), \delta\right)$-differential privacy, where $\varepsilon(\delta) \le \rho + 2\sqrt{\rho\log(1/\delta)}$.
\label{thm:privacy}
\end{theorem}

\noindent\textit{Proof sketch.} 
Fix $T\geq 1$ and $\alpha \in (0,1)$.
(i) At each iteration $t\in[T]$, \adaptive runs the exponential mechanism $k$ times with parameter $2\alpha\varepsilon_0$, which satisfies $\frac{k}{8}\roundbrack{2\alpha\varepsilon_0}^2 = \frac{k}{2}\roundbrack{\alpha\varepsilon_0}^2$-\textit{zCDP} \citep{cesar2020unifying}, and the Gaussian mechanism $k$ times with parameter $(1 - \alpha)\varepsilon_0$, which satisfies $ \frac{k}{2}\squarebrack{(1 - \alpha)\varepsilon_0}^2$-\textit{zCDP} \citep{BunS16}. 
(ii) using the composition theorem for concentrated differential privacy \citep{BunS16}, \adaptive satisfies $\frac{kT}{2}\squarebrack{\alpha^2 + (1-\alpha)^2}\varepsilon_0^2$-\textit{zCDP}  after $T$ iterations. 
(iii) Setting $\varepsilon_0 = \sqrt{\frac{2\rho}{kT \roundbrack{\alpha^2 + \roundbrack{1-\alpha}^2}}}$, we conclude that \adaptive satisfies $\rho$-\textit{zCDP}, which in turn implies $\roundbrack{\rho + 2\sqrt{\rho\log(1/\delta)}, \delta}$-\textit{differential privacy} for all $\delta > 0$ \citep{BunS16}.

%% file: docs/methods/pep.tex
\section{Maximum-Entropy Projection Algorithm}
Next, we propose \pep (Private Entropy Projection) under the framework \adaptive. Similar to \mwem, \pep employs the \emph{maximum entropy} principle, which also recovers a synthetic data distribution in the exponential family. However, since \pep adaptively assigns weights to past queries and measurements, it has faster convergence and better accuracy than \mwem. \pep's loss function in \adaptive can be derived through a constrained maximum entropy optimization problem.  At each round $t$, given a set of selected queries $\Qtil_{1:t}$ and their noisy measurements $\Atil_{1:t}$, the constrained optimization requires that the synthetic data $D_t$ satisfies $\gamma$-accuracy with respect to the noisy  measurements for all the selected queries in $\Qtil_{1:t}$. Then among the set of feasible distributions that satisfy accuracy constraints, \pep selects the distribution with maximum entropy, leading to the following regularized constraint optimization problem:
%
\begin{align}\label{eq:opt}
    \text{minimize:}\quad &\sum_{x\in\cX} D(x) \log\pp{D(x)} \\
    \notag
    \text{subject to: }\quad & \forall_{i\in [t]}\quad \left|\atil_i - \qtil_i(
    D)\right | \leq \gamma,
    \quad \sum_{x\in \cX} D(x) = 1
\end{align}
We can ignore the constraint that $\forall_{x\in \cX}  D(x) \geq 0, $ because it will be satisfied automatically. 

The solution of \eqref{eq:opt} is an exponentially weighted distribution parameterized by the dual variables $\lambda_1, \ldots, \lambda_t$ corresponding to the $t$ constraints. Therefore, if we solve the dual problem  of \eqref{eq:opt} in terms of the dual variables $\lambda_1, \ldots, \lambda_t$, then the distribution  that minimizes \eqref{eq:opt} is given by $D_t(x) \propto \exp\pp{\sum_{i=1}^t\lambda_i \qtil_i(x)}$. (Note that \mwem simply sets $\lambda_i = \tilde a_i - \tilde q_i(D_{t-1})$ without any optimization.) Given that the set of distributions is parameterized by the variables $\lambda_1, \ldots , \lambda_t$, the constrained optimization is then equivalent to minimizing the following exponential loss function:
\begin{align*}
    \cL^{\pep}\pp{\lambda, \Qtil_{1:t}, \Atil_{1:t}}
    =
    \log\pp{
\sum_{x\in \cX} \exp\pp{\sum_{i=1}^t \lambda_i \pp{\qtil_i(x) - \atil_i}}
}
+ \gamma \|\lambda\|_1
\end{align*}

Since \pep requires solving this constrained optimization problem on each round, we give an efficient iterative algorithm for solving \eqref{eq:opt}. We defer the details of the algorithm to the Appendix \ref{appx:pep}.

%% file: docs/methods/gem.tex
 \section{Overcoming Computational Intractability with Generative Networks}\label{sec:gem}



We introduce \gem (Generative Networks with the Exponential Mechanism), which optimizes over past queries to improve accuracy by training a generator network $G_\theta$ to implicitly learn a distribution of the data domain, where $G_\theta$ can be any neural network parametrized by weights $\theta$. As a result, our method \gem can compactly represent a distribution for any data domain while enabling fast, gradient-based optimization via auto-differentiation frameworks~\citep{NEURIPS2019_9015, tensorflow2015-whitepaper}.

Concretely, $G_\theta$ takes random Gaussian noise vectors $z$ as input and outputs a representation $G_\theta(z)$ of a product distribution over the data domain. Specifically, this product distribution representation takes the form of a $d'$-dimensional probability vector $G_\theta(z) \in [0, 1]^{d'}$, where $d'$ is the dimension of the data in one-hot encoding and each coordinate $G_\theta(z)_j$ corresponds to the marginal probability of a categorical variable taking on a specific value. To obtain this probability vector, we choose softmax as the activation function for the output layer in $G_\theta$.  
Therefore, for any fixed weights $\theta$, $G_\theta$ defines a distribution over $\cX$ through the generative process that draws a random $z\sim \mathcal{N}(0, \sigma^2 I)$ and then outputs random $x$ drawn from the product distribution $G_\theta(z)$. We will denote this distribution as $P_\theta$.

To define the loss function for \gem, we require that it be differentiable so that we can use gradient-based methods to optimize $G_\theta$. Therefore, we need to obtain a differentiable variant of $q$. Recall first that a query is defined by some predicate function $\phi:\cX \rightarrow\{0,1\}$ over the data domain  $\cX$ that evaluates over a single row $x \in \cX$. We observe then that one can extend any statistical query $q$ to be a function that maps a distribution $P_\theta$ over $\cX$ to a value in $[0,1]$:
\begin{equation}\label{eq:ptheta}
    q(P_\theta)
    = \mathbb{E}_{x\sim P_\theta}\squarebrack{\phi(x)} 
    =\sum_{x\in\cX} \phi(x) P_\theta(x)
\end{equation}
Note that any statistical query $q$ is then differentiable w.r.t. $\theta$:
\begin{align*}
    \nabla_\theta \left[ q(P_\theta) \right] 
    = \sum_{x \in \cX} \nabla_\theta P_\theta(x) \phi(x) 
    = \mathbb{E}_{\mathbf{z} \sim N(0, I_k)} \left[ \sum_{x \in \cX} \phi(x) \nabla_\theta \left[ \frac{1}{k} \sum_i^k \prod_{j=1}^{d'} (G_\theta(z_i)_j)^{x_j} \right] \right]
\end{align*}
and we can compute stochastic gradients of $q$ w.r.t. $\theta$ with random noise samples $z$. This also allows us to derive a differentiable loss function in the \adaptive framework. In each round $t$, given a set of selected queries $\Qtil_{1:t}$ and their noisy measurements $\Atil_{1:t}$, \gem minimizes the following $\ell_1$-loss:
\begin{equation}\label{eq:gemlossI}
    \cL^{\gem}\pp{\theta, \Qtil_{1:t}, \Atil_{1:t}}
    = \sum_{i=1}^t \abs{  
    {
       \qtil_i(P_\theta)
    } - \atil_{i}}.
\end{equation}
where $\qtil_i \in \Qtil_{1:t}$ and $\atil_i \in \Atil_{1:t}$.


In general, we can optimize $\cL^{\gem}$ by running stochastic (sub)-gradient descent. However, we remark that gradient computation can be expensive since obtaining a low-variance gradient estimate often requires calculating $\nabla_\theta P_\theta(x)$ for a large number of $x$. In Appendix \ref{appx:gem}, we include the stochastic gradient derivation for \gem and briefly discuss how an alternative approach from reinforcement learning.

For many query classes, however, there exists some closed-form, differentiable function surrogate to \eqref{eq:gemlossI} that evaluates $q(G_\theta(z))$ directly without operating over all $x \in \cX$. Concretely, we say that for certain query classes, there exists some representation $f_q:\Delta(\cX) \rightarrow [0,1]$ for $q$ that operates in the probability space of $\cX$ and is also differentiable. 

In this work, we implement \gem to answer $k$-way marginal queries, which have been one of the most important query classes for the query release literature \cite{hardt2010simple, vietri2020new, gaboardi2014dual, liu2021leveraging} and provides a differentiable form when extended to be a function over distributions. In particular, we show that $k$-way marginals can be rewritten as product queries (which are differentiable).
\begin{definition}[Product query]
Let $p\in\mathbb{R}^{d'}$ be a representation of a dataset (in the one-hot encoded space), and let $S \subseteq [d'] $ be some subset of dimensions of $p$. Then we define a product query $f_{S}$ as
\begin{equation}\label{eq:productquery}
    f_{S}(p) = \prod_{j\in S} p_j
\end{equation}
\end{definition}

A $k$-way marginal query $\phi$ can then be rewritten as \eqref{eq:productquery}, where $p = G_{\theta}(z)$ and $S$ is the subset of dimensions corresponding to the attributes $A$ and target values $y$ that are specified by $\phi$ (Definition \ref{def:marginals}). Thus, we can write any marginal query as $\prod_{j \in S} G_{\theta}(z)_j$, which is differentiable w.r.t. $G_\theta$ (and therefore differentiable w.r.t weights $\theta$ by chain rule). Gradient-based optimization techniques can then be used to solve \eqref{eq:gemlossI}; the exact details of our implementation can be found in Appendix \ref{appx:gem}.

%% file: docs/methods/methods_public.tex
\section{Extending to the \textit{public-data-assisted} setting}

Incorporating prior information from public data has shown to be a promising avenue for private query release \cite{bassily2020private, liu2021leveraging}. Therefore, we extend \gem to the problem of \textit{public-data-assisted private (PAP)} query release~\citep{bassily2020private} in which differentially private algorithms have access to public data. Concretely, we adapt \gem to utilize public data by initializing $D_0$ to a distribution over the public dataset. However, because in \gem, we implicitly model any given distribution using a generator $G$, we must first train without privacy (i.e., without using the exponential and Gaussian mechanisms) a generator $G_{\textrm{pub}}$, to minimize the $\ell_1$-error over some set of queries $\hat{Q}$. Note that in most cases, we can simply let $\hat{Q} = Q$ where $Q$ is the collection of statistical queries we wish to answer privately. \gempub then initializes $G_0$ to $G_{\textrm{pub}}$ and proceeds with the rest of the \gem algorithm.

\subsection{Overcoming limitations of \pmwpub.} \label{sec:pub_limitations}

We describe the limitations of \pmwpub by providing two example categories of public data that it fails to use effectively. We then describe how \gempub overcomes such limitations in both scenarios.
 
\paragraph{Public data with insufficient support.} 
We first discuss the case in which the public dataset has an insufficient support, which in this context means the support has high \textit{best-mixture-error} \citep{liu2021leveraging}. Given some support $S\subseteq \cX$, the \textit{best-mixture-error} can be defined as
\begin{align*}
    \min_{\mu \in \Delta(S)} 
    \max_{q\in Q}
    \left|
    q\pp{D} - 
    \sum_{x \in S} \mu_x  q(x)
    \right|
\end{align*}
where $\mu \in \Delta(S)$ is a distribution over the set $S$ with $\mu(x) \ge 0$ for all $x\in S$ and $\sum_{x\in S} \mu(x) = 1$.

In other words, the \textit{best-mixture-error} approximates the lowest possible max error that can be achieved by reweighting some support, which in this case means \pmwpub cannot achieve max errors lower that this value. While \citet{liu2021leveraging} offer a solution for filtering out poor public datasets ahead of time using a small portion of the privacy budget, \pmwpub cannot be run effectively if no other suitable public datasets exist. \gempub however avoids this issue altogether because unlike \mwem (and therefore \pmwpub), which cannot be run without restricting the size of $\cD$, \gempub can utilize public data without restricting the distributional family it can represent (since both \gem and \gempub compactly parametrize any distribution using a neural network).

\paragraph{Public data with incomplete data domains.} Next we consider the case in which the public dataset only has data for a subset of the attributes found in the private dataset. We note that as presented in \citet{liu2021leveraging}, \pmwpub cannot handle this scenario. One possible solution is to augment the public data distribution by assuming a uniform distribution over all remaining attributes missing in the public dataset. However, while this option may work in cases where only a few attributes are missing, the missing support grows exponentially in the dimension of the missing attributes. In contrast, \gempub can still make use of such public data. In particular, we can pretrain a generator $G$ on queries over just the attributes found in the public dataset. Again, \gempub avoids the computational intractability of \pmwpub in this setting since it parametrizes its output distribution with $G$.

%% file: docs/experiments/experiments.tex
\section{Empirical Evaluation} 
In this section, we empirically evaluate \gem and \pep against baseline methods on the ACS~\citep{ruggles2020ipums} and ADULT~\citep{Dua:2019} datasets in both the standard\footnote{We refer readers to Appendix \ref{appx:adult_loans} where we include an empirical evaluation on versions of the ADULT and LOANS~\citep{Dua:2019} datasets used in other related private query release works \citep{mckenna2018optimizing, vietri2020new, aydore2021differentially}.} and \textit{public-data-assisted} settings.

\paragraph{Data.}
To evaluate our methods, we construct public and private datasets from the ACS and ADULT datasets by following the preprocessing steps outlined in \citet{liu2021leveraging}. For the ACS, we use 2018 data for the state of Pennsylvania (PA-18) as the private dataset. For the public dataset, we select 2010 data for Pennsylvania (PA-10) and 2018 data for California (CA-18). In our experiments on the ADULT dataset, private and public datasets are sampled from the complete dataset (using a 90-10 split). In addition, we construct low-dimensional versions of both datasets, which we denote as ACS (reduced) and ADULT (reduced), in order to evaluate \pep and \mwem.

\paragraph{Baselines.}
We compare our algorithms to the strongest performing baselines in both low and high-dimensional settings, presenting results for \mwem, \dq, and \rapsoftmax in the standard setting\footnote{Having consulted \citet{mckenna2018optimizing}, we concluded that running \hdmm is infeasible for our experiments, since it generally cannot handle a data domain with size larger than $10^9$. See Appendix \ref{appx:hdmm} for more details.}\footnote{Because \rap performs poorly relative to the other methods in our experiments, plotting its performance would make visually comparing the other methods difficult. Thus, we exclude it from Figure \ref{fig:max_error_acs_adult} and refer readers to Appendix \ref{appx:rap_compare}, where we present failure cases for \rap and compare it to \rapsoftmax.} and \pmwpub in the \textit{public-data-assisted} setting.

\paragraph{Experimental details.}
To present a fair comparison, we implement all algorithms using the privacy mechanisms and \textit{zCDP} composition described in Section \ref{sec:privacy}. To implement \gem for $k$-way marginals, we select a simple multilayer perceptron for $G_\theta$. Our implementations of \mwem and \pmwpub output the last iterate $D_t$ instead of the average and apply the multiplicative weights update rule using past queries according to the pseudocode described in \citet{liu2021leveraging}. We report the best performing $5$-run average across hyperparameter choices (see Tables \ref{tab:hyperparameters_pep}, \ref{tab:hyperparameters_gem}, \ref{tab:hyperparameters_pep_pub}, \ref{tab:hyperparameters_gem_pub}, and \ref{tab:hyperparameters_baselines} in Appendix \ref{appx:experiments}) for each algorithm.

\input{docs/experiments/results}

%% file: docs/experiments/results.tex
\input{figures/max_error/acs_adult}

\input{figures/max_error/pub}

\subsection{Results}\label{sec:results}

\textbf{Standard setting.} In Figure \ref{fig:max_error_acs_adult}, we observe that in low-dimensional settings, \pep and \gem consistently achieve strong performance compared to the baseline methods. While \mwem and \pep are similar in nature, \pep outperforms \mwem on both datasets across all privacy budgets except $\varepsilon \in \curlybrack{0.1, 0.15}$ on ACS (reduced), where the two algorithms perform similarly. In addition, both \pep and \gem outperform \rapsoftmax. Moving on to the more realistic setting in which the data dimension is high, we again observe that \gem outperforms \rapsoftmax on both datasets.


\paragraph{\textit{Public-data-assisted} setting.} To evaluate the query release algorithm in the public-data-assisted setting, we present the three following categories of public data:

\textit{Public data with sufficient support.} To evaluate our methods when the public dataset for ACS PA-18 has low \textit{best-mixture-error}, we consider the public dataset ACS PA-10. We observe in Figure \ref{fig:max_error_acs_adult} that \gempub performs similarly to \pmwpub, with both outperforming \gem (without public data).

\textit{Public data with insufficient support.} In Figure \ref{fig:max_error_pub}a, we present CA-18 as an example of this failure case in which the \textit{best-mixture-error} is over $10\%$, and so for any privacy budget, \pmwpub cannot achieve max errors lower that this value. However, for the reasons described in Section \ref{sec:pub_limitations}, \gempub is not restricted by \textit{best-mixture-error} and significantly outperforms \gem (without public data) when using either public dataset.

\textit{Public data with incomplete data domains.} To simulate this setting, we construct a reduced version of the public dataset in which we keep only $7$ out of $13$ attributes in ADULT. In this case, $6$ attributes are missing, and so assuming a uniform distribution over the missing attributes would cause the dimension of the approximating distribution $D$ to grow from $\approx 4.4 \times 10^3$ to a $\approx 3.2 \times 10^9$. \pmwpub would be computationally infeasible to run in this case. To evaluate \gempub, we pretrain the generator $G$ using all $3$-way marginals on both the complete and reduced versions of the public dataset and then finetune on the private dataset (we denote these two finetuned networks as \gempub and \gempubred respectively). We present results in Figure \ref{fig:max_error_pub}b. Given that the public and private datasets are sampled from the same distribution, \gempub unsurprisingly performs extremely well. However, despite only being pretrained on a small fraction of all $3$-way marginal queries ($\approx 20k$ out $334k$), \gempubred is still able to improve upon the performance of \gem and achieve lower max error for all privacy budgets.

%% file: figures/max_error/acs_adult.tex
\begin{figure}[!t]
    \centering
    \includegraphics[width=\linewidth]{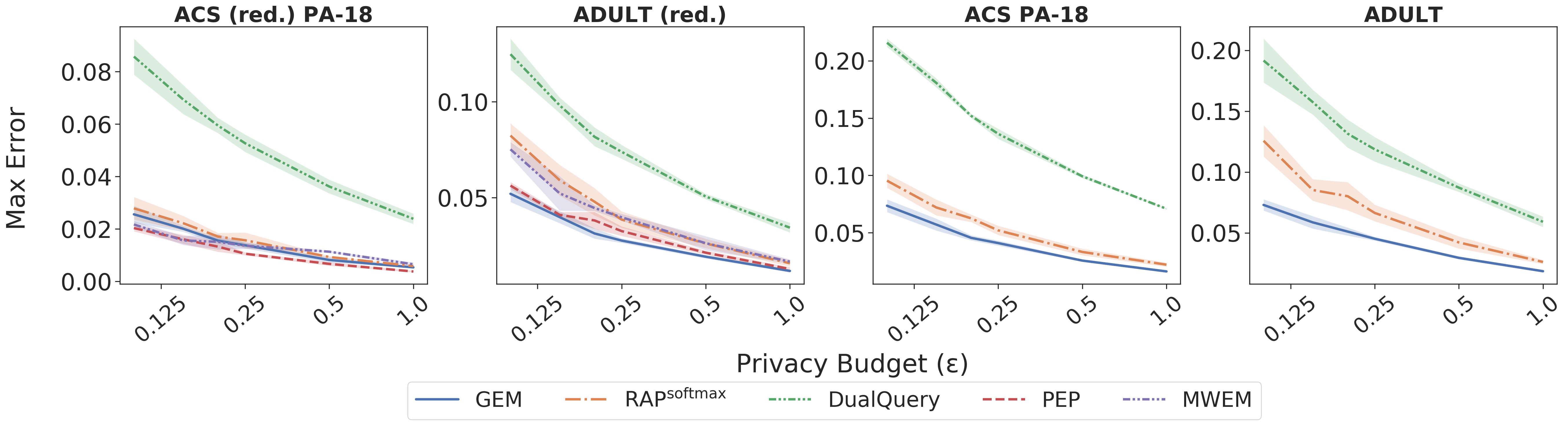}
    \caption{
    Max error for $3$-way marginals evaluated on ADULT and ACS PA-18 using privacy budgets $\varepsilon \in \{ 0.1, 0.15, 0.2, 0.25, 0.5, 1 \}$ and $\delta = \frac{1}{n^2}$. The \textit{x-axis} uses a logarithmic scale. We evaluate using the following workload sizes:
    ACS (reduced) PA-18: $455$;
    ADULT (reduced): $35$; 
    ACS PA-18: $4096$; 
    ADULT: $286$.
    Results are averaged over $5$ runs, and error bars represent one standard error.
    }
    \label{fig:max_error_acs_adult}
\end{figure}

%% file: figures/max_error/pub.tex
\begin{figure}[!t]
    \centering
    \includegraphics[width=\linewidth]{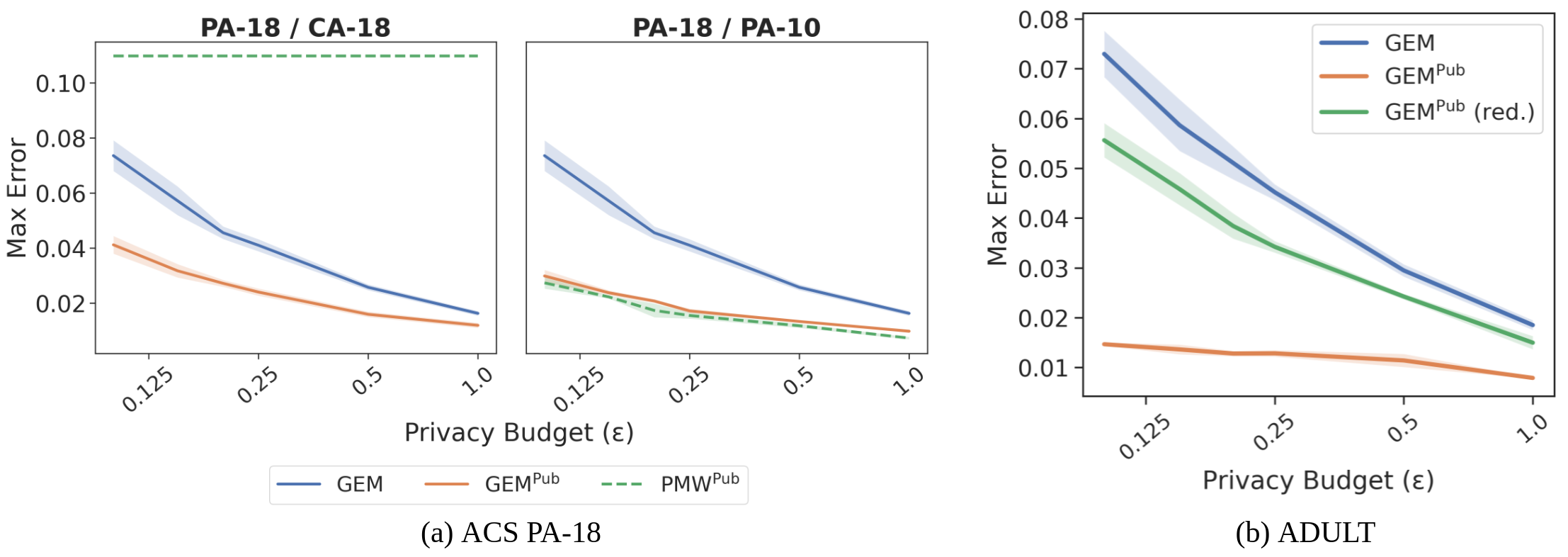}
    \caption{
    Max error for $3$-way marginals with privacy budgets $\varepsilon \in \{ 0.1, 0.15, 0.2, 0.25, 0.5, 1 \}$ and $\delta = \frac{1}{n^2}$. The \textit{x-axis} uses a logarithmic scale. Results are averaged over $5$ runs, and error bars represent one standard error. 
    \textbf{(a)} ACS PA-18 (workloads = $4096$). We evaluate public-data-assisted algorithms with the following public datasets:
    \textit{Left:} 2018 California (CA-18);
    \textit{Right:} 2010 Pennsylvania (PA-10).
    \textbf{(b)} ADULT (workloads = $286$). We evaluate \gem using both the complete public data (\gempub) and a reduced version that has fewer attributes (\gempubred) .
    }
    \label{fig:max_error_pub}
\end{figure}

%% file: docs/appendix/appendix.tex
\appendix

\input{docs/appendix/adaptive}

\newpage

\input{docs/appendix/gem}

\newpage

\input{docs/appendix/pep}

\newpage

\input{docs/appendix/experiments}

%% file: docs/appendix/adaptive.tex
\section{\adaptive}\label{appx:adaptive}

\input{docs/appendix/marginal_trick}

\subsection{Choices of loss functions and distributional families}\label{appx:loss_functions}

We provide additional details about each iterative algorithm, including the loss function $\cL$ and and distributional family $\cD$ (under the \adaptive framework).


\paragraph{\mwem from \citet{hardt2010simple}} 
The traditional \mwem algorithm samples one query each round, where after $t$ rounds, the set of queries/measurements is $\Qtil_t = \{\qtil_1, \ldots, \qtil_t\}, \Atil_t=\{\atil_1, \ldots, \atil_t \}$. 
Let the $t-1$ previous solutions be $D_1, \ldots D_{t-1}$. Then \mwem solves an entropy regularized problem in which it finds $D_t$ that minimizes the following loss:  
\begin{align*}
    \cL^{\mwem}(D, \Qtil_{1:t}, \Atil_{1:t})
    = \sum_{i=1}^t \sum_{x\in \cX} D(x) \qtil_i(x) \pp{\atil_i - \qtil_i(D_{i-1})}
    + \sum_{x\in\cX} D(x) \log D(x)
\end{align*}
We can show that if $D_t = \argmin_{D\in \Delta(\cX) } \cL^{\mwem}(D, \Qtil_t, \Atil_t)$  then $D_t$ evaluates to $D_t(x) \propto \exp\pp{-\sum_{i=1}^{t} \qtil_i(x)(\atil_i - \qtil_i(D_{i-1})) }$ which is the exactly the distribution computed by \mwem. See \ref{sec:mwemupdate} for derivation. We note that \mwem explicitly maintains (and outputs) a distribution $D \in \cD$ where $\cD$ includes all distributions over the data domain $\cX$, making it computationally intractable for high-dimension settings.

\paragraph{\dq from \citet{gaboardi2014dual}}
\dq is a special case of the \adaptive framework in which the measurement step is skipped (abusing notation, we say $\alpha=1$). Over all iterations of the algorithm,  \dq keeps track of a probability distribution over the set of queries $\cQ$ via multiplicative weights, which we denote here by $\cQ_t\in \Delta(Q)$. 
On round $t$, \dq samples $s$ queries ($\Qtil_t = \{\qtil_{t,1}, \ldots \qtil_{t,s}\}$) from $\cQ_t$ and outputs $D_t$ that minimizes the the following loss function:
\begin{align*}
    \cL^{\dq}(D, \Qtil_t)
    = \sum_{i=1}^{s}  \qtil_{t,i}(D)
\end{align*}
The optimization problem for $\cL^{\dq}(D, \Qtil_t)$ is NP-hard. However,  
the algorithm encodes the problem as a mix-integer-program (MIP) and takes advantage of available fast solvers. The final output of \dq is the average $\frac{1}{T}\sum_{t=1}^T D_t$, which we note implicitly describes some empirical distribution over $\cX$.

\paragraph{\fem from \citet{vietri2020new}}
The algorithm \fem follows a follow the perturbed leader strategy. 
As with \mwem, the algorithm \fem samples one query each round using the exponential mechanism, so that the set of queries in round $t$ is $\Qtil_t = \{\qtil_1, \ldots, \qtil_t\}$. Then on round $t$, \fem chooses the next distribution by solving:
\begin{align*}
  \cL^{\fem}(D, \Qtil_{1:t}) = 
    \sum_{i=1}^t \qtil_t(D) + \mathbb{E}_{x\sim D, \eta\sim \text{Exp}(\sigma)^d}\pp{\langle x, \eta \rangle} 
\end{align*}

Similar to \dq, the optimization problem for $\cL^{\fem}$ also involves solving an NP-hard problem. Additionally, because the function $\cL^{\fem}$ does not have a closed form due to the expectation term, \fem follows a sampling strategy to approximate the optimal solution. 
On each round, $\fem$ generates $s$ samples, where each sample is obtained in the following way:  
Sample a noise vector $\eta \sim \text{Exp}(\sigma)^d$ from the exponential distribution and use a MIP to solve $x_{t,i}\leftarrow \argmin_{x\in \cX} \sum_{i=1}^t \qtil_t(D)+\langle x, \eta \rangle $ for all $i\in[s]$. Finally, the output on round $t$ is the empirical distribution derived from the $s$ samples: $D_{t} = \{x_{t,1}, \ldots, x_{t,s}\}$.
The final output is the average $\frac{1}{T} \sum_{t=1}^T D_{t}$.

\paragraph{\rapsoftmax adapted from \citet{aydore2021differentially}}
At iteration $t$, \rapsoftmax solves the following optimization problem:
\begin{align*}
    \cL^{\rap}(D, \Qtil_{1:t}, \Atil_{1:t})
    = \sum_{i, j} \roundbrack{\qtil_{i,j}(D) - \atil_{i,j}}^2
\end{align*}
As stated in Section \ref{appx:adaptive}, we apply the softmax function such that \rapsoftmax outputs a synthetic dataset drawn from some probabilistic family of distributions $\cD = \curlybrack{\sigma(M) | M \in \mathbb{R}^{n' \times d}}$.

\input{docs/appendix/mwem_update}

%% file: docs/appendix/marginal_trick.tex
\subsection{$k$-way marginals sensitivity}\label{appx:marginal_trick}

Typically, iterative private query release algorithms assume that the query class $Q$ contains sensitivity $1$ queries \citep{hardt2010multiplicative, gaboardi2014dual, vietri2020new, aydore2021differentially, liu2021leveraging}. However, recall that a $k$-way marginal query is defined by a subset of features $S \subseteq [d]$ and a target value $y\in \prod_{i\in S} \cX_i$ (Definition \ref{def:marginals}). Given a feature set $S$ (with $|S|= k$), we can define a workload $W_S$ as the set of queries defined over the features in $S$. 
\begin{align*}
    W_S = \left\{ q_{S,y} : y \in \prod_{i\in S} \cX_i \right\}
\end{align*}

Then for any dataset $D$, the workload's answer is given by $W_S(D) = \{q_{S,y} (D)\}_{y \in \prod_{i\in S}\cX_i}$, where $W_S$ has $\ell_2$-sensitivity equal to $\sqrt{2}$.
Therefore to achieve more efficient privacy accounting for $k$-way marginals in \adaptive, we can use the exponential mechanism to select an entire workload $\hat{W}_S$ that contains the max error query and then obtain measurements for \emph{all} queries in $\hat{W}_S$ using the Gaussian mechanism, adding noise
\begin{align*}
    z \sim \mathcal{N} \roundbrack{0, \roundbrack{\frac{\sqrt{2}}{n(1-\alpha)\varepsilon_0}}^2}.
\end{align*}
to each query in $W_S$.

In appendix \ref{appx:experiments_marginal_trick}, we show that using this marginal trick significantly improves the performance of \gem and therefore recommend this type of privacy accounting when designing query release algorithms for $k$-way marginal queries. 

%% file: docs/appendix/mwem_update.tex
\subsection{\mwem update} \label{sec:mwemupdate}

Given the loss function: 
\begin{equation}\label{eq:mwemloss}
    \cL^{\text{\mwem}}(D, \Qtil_t, \Atil_t)
    = \sum_{i=1}^t \sum_{x\in \cX} D(x) \qtil_i(x) \pp{\atil_i - \qtil_i(D_{i-1})}
    + \sum_{x\in\cX} D(x) \log\pp{D(x)}
\end{equation}

The optimization problem becomes $D_t = \argmin_{D\in \Delta(\cX) } \cL^{\text{mwem}}(D, \Qtil_t, \Atil_t)$. 
The solution $D$ is some distribution, which we can express as a constraint $\sum_{x\in \cX} D(x) = 1$. Therefore, this problem is a constrained optimization problem.
To show that \eqref{eq:mwemloss} is the \mwem's true loss function,
we can write down the Lagrangian as:
\begin{align*}
        \cL
    = \sum_{i=1}^t \sum_{x\in \cX} D(x) \qtil_i(x) \pp{\atil_i - \qtil_i(D_{i-1})}
    + \sum_{x\in\cX} D(x) \log\pp{D(x)}
    + \lambda \pp{\sum_{x\in \cX} D(x) - 1}
\end{align*}

Taking partial derivative with respect to $D(x)$:
\begin{align*}
      \frac{\partial \cL}{\partial D(x) }  
    = \sum_{i=1}^{t}  \qtil_i(x) \pp{\atil_i - \qtil_i(D_{i-1})}
    + (1 + \log D(x) ) 
    + \lambda
\end{align*}
Setting $ \frac{\partial \cL}{\partial D(x) } = 0$ and solving for $D(x)$:
\begin{align*}
D(x) = 
\exp\pp{-1 - \lambda -\sum_{i=1}^{t}  \qtil_i(x) \pp{\atil_i - \qtil_i(D_{i-1})}  }
\end{align*}

Finally, the value of $\lambda $ is set such that $D$ is a probability distribution:

\begin{align*}
D(x) = 
\frac{
\exp\pp{-\sum_{i=1}^{t}  \qtil_i(x) \pp{\atil_i - \qtil_i(D_{i-1})}  }
}
{
\sum_{x\in\cX}\exp\pp{ -\sum_{i=1}^{t}  \qtil_i(x) \pp{\atil_i - \qtil_i(D_{i-1})}}
}
\end{align*}
This concludes the derivation of \mwem loss function.

%% file: docs/appendix/gem.tex
\section{\gem}\label{appx:gem}

We show the exact details of \gem in Algorithms \ref{alg:gem} and \ref{alg:gemupdate}. Note that given a vector of queries $Q_t = \anglebrack{q_1, \ldots, q_t}$, we define $f_{Q_t}(\cdot) = \anglebrack{f_{q_1}(\cdot), \ldots, f_{q_t}(\cdot)}$.

\input{docs/algos/gem}

\input{docs/algos/gemupdate}

\subsection{Loss function (for $k$-way marginals) and distributional family}

For any $z \in \mathbb{R}$, $G(z)$ outputs a distribution over each attribute, which we can use to calculate the answer to a query via $f_q$. In \gem however, we instead sample a noise vector $\vect{z} = \anglebrack{z_1 \ldots z_B}$ and calculate the answer to some query $q$ as $\frac{1}{B} \sum_{j=1}^{B} f_{q} \roundbrack{G\roundbrack{z_{j}}}$. One way of interpreting the batch size $B$ is to consider each $G\roundbrack{z_{j}}$ as a unique distribution. In this sense, \gem models $B$ sub-populations that together comprise the overall population of the synthetic dataset. Empirically, we find that our model tends to better capture the distribution of the overall private dataset in this way (Figure \ref{fig:all_errors_gem_batch_size}). Note that for our experiments, we choose $B=1000$ since it performs well while still achieving good running time. However, this hyperparameter can likely be further increased or tuned (which we leave to future work).

\input{figures/all_errors/gem_batch_size}

Therefore, using this notation, \gem outputs then a generator $G \in \cD$ by optimizing $\ell_1$-loss at each step $t$ of the \adaptive framework:
\begin{equation}\label{eq:gemloss_kway}
    \cL^{\gem}\pp{G, \Qtil_{1:t}, \Atil_{1:t}}
    = \sum_{i=1}^t \abs{\frac{1}{B} \sum_{j=1}^{B} f_{\qtil_i} \roundbrack{G\roundbrack{z_{j}}} - \atil_{i}}
\end{equation}

Lastly, we note that we can characterize family of distributions $\cD = \curlybrack{G_\theta(z) | z \sim \mathcal{N}(0, 1)}$ in \gem by considering it as a class of distributions whose marginal densities are parameterized by $\theta$ and Gaussian noise $z \sim \mathcal{N}(0, 1)$ We remark that such densities can technically be characterized as Boltzmann distributions.

\subsection{Additional implementation details}

\paragraph{\ema\ output}
We observe empirically that the performance of the last generator $G_T$ is often unstable. One possible solution explored previously in the context of privately trained GANs is to output a mixture of samples from a set of generators \cite{beaulieu2019privacy, neunhoeffer2020private}. In our algorithm \gem, we instead draw inspiration from \citet{yazici2018unusual} and output a single generator $G_{out}$ whose weights $\theta_{out}$ are an exponential moving average (\ema) of weights $\theta_t$ obtained from the latter half of training. More concretely, we define $\theta_{out} = \ema \roundbrack{\curlybrack{\theta_j}_{j=\frac{T}{2}}^T}$, where the update rule for \ema\ is given by $\theta^{EMA}_{k}=\beta \theta^{EMA}_{k-1}+(1-\beta) \theta_{k}$ for some parameter $\beta$.

\paragraph{Stopping threshold $\gamma$}
To reduce runtime and prevent \gem from overfitting to the sampled queries, we run \gemupdate with some early stopping threshold set to an error tolerance $\gamma$. Empirically, we find that setting $\gamma$ to be half of the max error at each time step $t$. Because sampling the max query using the exponential mechanism provides a noisy approximation of the true max error, we find that using an exponential moving average (with $\beta=0.5$) of the sampled max errors is a more stable approximation of the true max error. More succinctly, we set $\gamma = \ema(\curlybrack{c_i}_{i=0}^{t})$ where $c_i$ is max error at the beginning of iteration $i$.

\paragraph{Resampling Gaussian noise.}
In our presentation of \gem and in Algorithms \ref{alg:gem} and \ref{alg:gemupdate}, we assume that \gem resamples Gaussian noise $z$. While resampling $z$ encourages \gem to train a generator to output a distribution for any $\vect{z} \sim \mathcal{N}(0, I_B)$ for some fixed batch size $B$, we find that fixing the noise vector $\vect{z}$ at the beginning of training leads to faster convergence. Moreover in Figure \ref{fig:all_errors_gem_resample}, we show that empirically, the performance between whether we resample $z$ at each iteration is not very different. Since resampling $z$ does not induce any benefits to generating synthetic data for the purpose of query release, in which the goal is output a single synthetic dataset or distribution, we run all experiments without resampling $z$. However, we note that it is possible that in other settings, resampling the noise vector at each step makes more sense and warrants compromising per epoch convergence speed and overall runtime. We leave further investigation to future work. 

\input{figures/all_errors/gem_resample}

\subsection{Optimizing over arbitrary query classes}

To optimize the loss function for \gem (Equation \ref{eq:gemlossI}) using gradient-based optimization, we need to have access to the gradient of each $\qtil_i$ with respect to the input distribution $G_\theta(z)$ for any $z$ (once we compute this gradient, we can then derive the gradient of the loss function with respect to the parameters $\theta$ via chain rule). Given any arbitrary query function $q$, we can rewrite it as \eqref{eq:ptheta}, which is differentiable w.r.t. $\theta$. 

More specifically, for $\mathbf{z} \sim N(0, I_k)$, we write $P_\theta (x) = \mathbb{E}_{\mathbf{z} \sim N(0, I_k)}\left[ P_{\theta, \mathbf{z}} (x) \right]$, where
\begin{align*}
    P_{\theta, \mathbf{z}} (x) = \frac{1}{k} \sum_{i=1}^k \prod_{j=1}^{d'} (G_\theta(z_i)_j)^{x_j}
\end{align*}

Then
\begin{align*}
    \nabla_\theta [q(P_\theta)]
    &= \nabla_\theta \sum_{x \in \cX} \phi(x)  P_\theta(x) \\
    &= \sum_{x \in \cX} \phi(x) \nabla_\theta P_\theta(x) \\ 
    &= \sum_{x \in \cX} \phi(x) \nabla_\theta \left[ \mathbb{E}_{\mathbf{z} \sim N(0, I_k)}\left[ P_{\theta, \mathbf{z}} (x) \right] \right] \\ 
    &= \mathbb{E}_{\mathbf{z} \sim N(0, I_k)}\left[ \sum_{x \in \cX} \phi(x) \nabla_\theta \left[ P_{\theta, \mathbf{z}} (x) \right] \right]
\end{align*}

However, while this form allows us to compute the gradient $\nabla_{\theta} q$ even when $\phi$ itself may not be differentiable w.r.t. $x$ or have a closed form, this method is not computationally feasible when the data domain $\cX$ is too large because it requires evaluating $\phi$ on all $x\in\cX$. One possible alternative is to construct an unbiased estimator. However, this estimator may suffer from high variance when the number of samples is insufficiently large, inducing a trade-off between computational efficiency and the variance of the estimator. 

Incorporating techniques from reinforcement learning, such as the REINFORCE algorithm \citep{williams1992simple}, may serve as alternative ways for optimizing over non-differentiable queries. Specifically, we can approximate \eqref{eq:ptheta} in the following way:

\begin{align*}
    \nabla_\theta [q(P_\theta)] 
    &= \mathbb{E}_{\mathbf{z} \sim N(0, I_k)}\left[ \sum_{x \in \cX} \phi(x) \nabla_\theta P_{\theta, \mathbf{z}} (x) \right] \\
    &= \mathbb{E}_{\mathbf{z} \sim N(0, I_k)}\left[ \sum_{x \in \cX} \phi(x) \frac{P_{\theta, \mathbf{z}}}{P_{\theta, \mathbf{z}}} \nabla_\theta P_{\theta, \mathbf{z}} (x) \right] \\ 
    &= \mathbb{E}_{\mathbf{z} \sim N(0, I_k)}\left[ \sum_{x \in \cX} \phi(x) P_{\theta, \mathbf{z}} (x) \nabla_\theta \log P_{\theta, \mathbf{z}} (x) \right]
\end{align*}

We can then approximate this gradient by drawing $m$ samples $\{ x_1 \ldots x_m \}$ from $P_{\theta, \mathbf{z}}(x)$, giving us
\begin{align*}
    \mathbb{E}_{\mathbf{z} \sim N(0, I_k)}\left[ \frac{1}{m} \sum_{i=1}^k \phi(x_i) \nabla_\theta \log P_{\theta, \mathbf{z}}(x_i) \right]
\end{align*}

Further work would be required to investigate whether optimizing such surrogate loss functions is effective when differentiable, closed-form representations of a given query class (e.g., the product query representation of $k$-way marginals) are unavailable. 

%% file: docs/algos/gem.tex
\begin{algorithm}[H]
\SetAlgoLined
\textbf{Input:} Private dataset $\Dpriv$, set of differentiable queries $Q$ \\
\textbf{Parameters}: privacy parameter $\rho$, number of iterations $T$, privacy weighting parameter $\alpha$, batch size $B$, stopping threshold $\gamma$ \\
Initialize generator network $G_0$ \\
Let $\varepsilon_0 = \sqrt{\frac{2\rho}{T \roundbrack{\alpha^2 + \roundbrack{1-\alpha}^2}}}$ \

 \For{$t = 1 \ldots T$}{
  \textbf{Sample:} 
  Sample $\vect{z} = \anglebrack{z_1 \ldots z_B} \sim \mathcal{N}(0, I_B)$ \\
  Choose $\qtil_t$ using the \textit{exponential mechanism} with score 
  $$\pr{q_t = q} \propto \exp\roundbrack{\frac{\alpha \varepsilon_0 n}{2} |q(\Dpriv) - q(G_{t-1}\roundbrack{\vect{z}})|}$$ \\
  
  \textbf{Measure:}
  Let $\atil_t = \qtil_t(\Dpriv) + \mathcal{N} \roundbrack{0, \roundbrack{\frac{1}{n(1-\alpha)\varepsilon_0}}^2}$ \\
  
  \textbf{Update:} 
  $G_t = \gemupdate \roundbrack{G_{t-1}, Q_{t}, \vect{\atil_t}, \gamma}$
  where $Q_{t} = \anglebrack{\qtil_1, \ldots, \qtil_t}$ and $\vect{\atil_t} = \anglebrack{\atil_1, \ldots, \atil_t}$ \\
 }
 Let $\theta_{out} = \ema \roundbrack{\curlybrack{\theta_j}_{j=\frac{T}{2}}^T}$ where $\theta_j$ parameterizes $G_j$ \\
 Let $G_{out}$ be the generator parameterized by $\theta_{out}$ \\
 Output $G_{out} \roundbrack{\vect{z}}$ \\
 \caption{\sf{GEM}}
 
 \label{alg:gem}
\end{algorithm}

%% file: docs/algos/gemupdate.tex
\begin{algorithm}[H]
\SetAlgoLined
\textbf{Input:} Generator $G$ parameterized by $\theta$, queries $Q$, noisy measurements $\vect{\atil}$, stopping threshold $\gamma$ \\
\textbf{Parameters:} max iterations $T_{\text{max}}$, batch size $B$ \\
Sample $\vect{z} = \anglebrack{z_1 \ldots z_B} \sim \mathcal{N}(0, I_B)$ \\
Let $\vect{c} = \vect{\atil} - \frac{1}{B} \sum_{j=1}^{B} f_{Q} \roundbrack{G\roundbrack{z_{j}}}$ be errors over queries $Q$ \\
Let $i = 0$  \\
\While{$i < T_{\text{max}}$ and $\lVert \vect{c} \rVert_\infty \ge \gamma$}{
  Let $J = \curlybrack{j \mid |c_{j}| \ge \gamma}$ \\
  Update $G$ to minimize the loss function with the stochastic gradient $
  \nabla_{\theta} \frac{1}{|J|}\sum_{j \in J}
  |c_{ij}|
  $ \\
  Sample $\vect{z} = \anglebrack{z_1 \ldots z_B} \sim \mathcal{N}(0, I_B)$ \\
  Let $\vect{c} = \vect{\atil} - \frac{1}{B} \sum_{j=1}^{B} f_{Q} \roundbrack{G\roundbrack{z_{j}}}$ \\
  Let $i = i + 1$ \\
 }
 \textbf{Output:} $G$ \\
 \caption{\gemupdate}
 \label{alg:gemupdate}
\end{algorithm}

%% file: figures/all_errors/gem_batch_size.tex
\begin{figure}[!t]
    \centering
    \includegraphics[width=\linewidth]{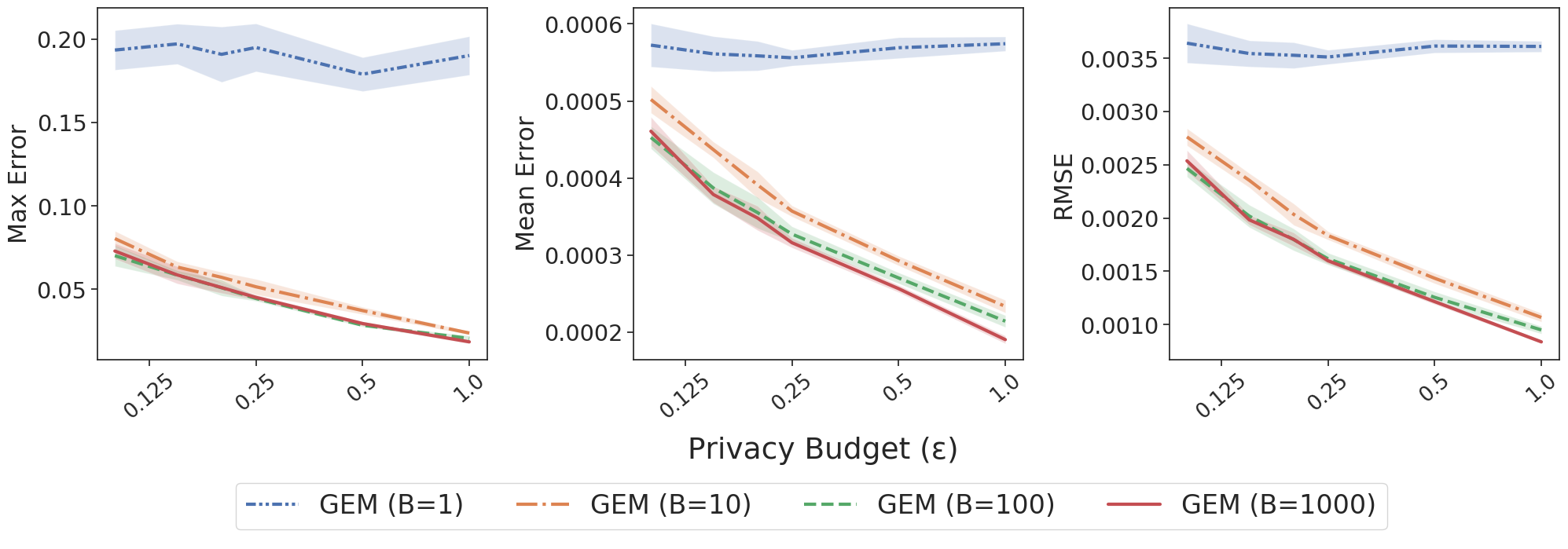}
    \caption{
    Error comparison of \gem using different batch sizes $B$ on ADULT (workloads=286), evaluated on $3$-way marginals with privacy budgets $\varepsilon \in \{ 0.1, 0.15, 0.2, 0.25, 0.5, 1 \}$ and $\delta = \frac{1}{n^2}$. The \textit{x-axis} uses a logarithmic scale. Results are averaged over $5$ runs, and error bars represent one standard error.
    }
    \label{fig:all_errors_gem_batch_size}
\end{figure}

%% file: figures/all_errors/gem_resample.tex
\begin{figure}[!t]
    \centering
    \includegraphics[width=\linewidth]{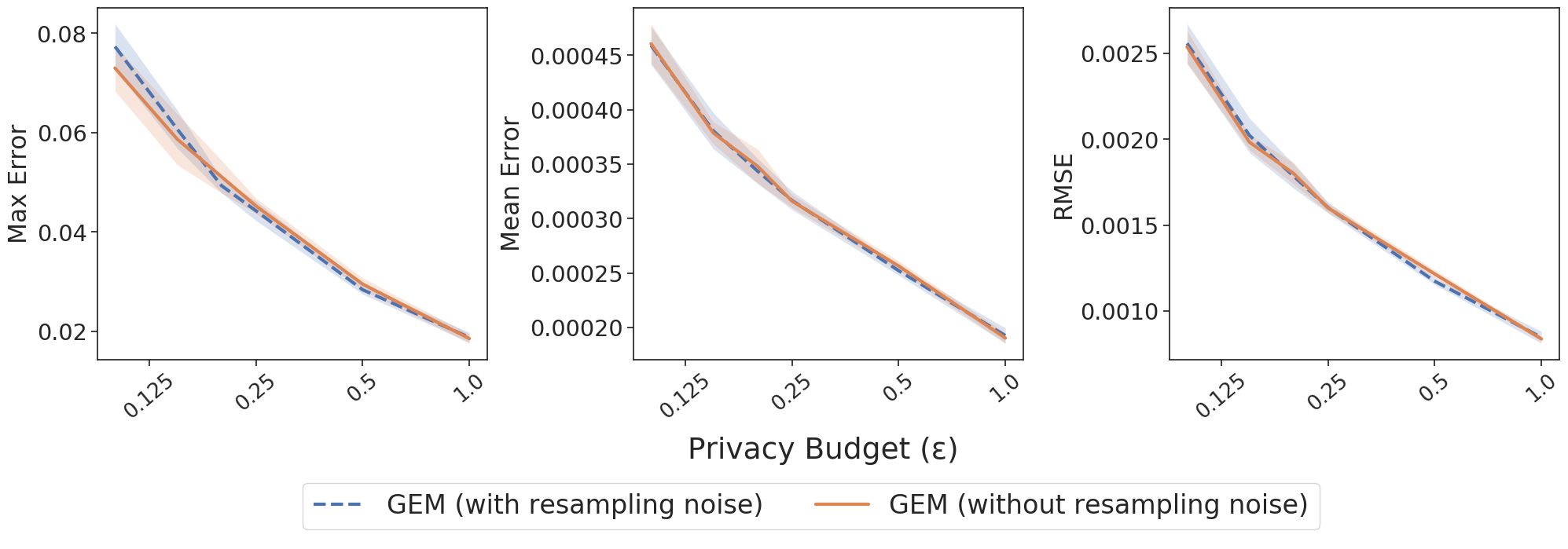}
    \caption{
    Error comparison of \gem with and without resampling $\vect{z}$ at each step on ADULT (workloads=286), evaluated on $3$-way marginals with privacy budgets $\varepsilon \in \{ 0.1, 0.15, 0.2, 0.25, 0.5, 1 \}$ and $\delta = \frac{1}{n^2}$. The \textit{x-axis} uses a logarithmic scale. Results are averaged over $5$ runs, and error bars represent one standard error.
    }
    \label{fig:all_errors_gem_resample}
\end{figure}

%% file: docs/appendix/pep.tex
\section{\pep}\label{appx:pep}

In the following two sections, we first derive \pep's loss function and and in the next section we derive \pep's update rule (or optimization procedure) that is used to minimize it's loss function.

\input{docs/appendix/pep_update}
\input{docs/appendix/pep_iter_oracle}

%% file: docs/appendix/pep_update.tex
\subsection{\pep loss function}

%
In this section we derive the loss function that algorithm \pep optimizes over on round $t$. Fixing round $t$, we let $\Qtil_t \subset Q$ be a subset of queries that were selected using a private mechanism and and let \textbf{$\Atil_t$} be the noisy measurements corresponding to $\Qtil_t$.
Then algorithm \pep finds a feasible solution to the problem:
\begin{align}\label{appex:eq:opt}
    \text{minimize:}\quad &\diver{D \parallel U} \\
    \notag
    \text{subject to: }\quad & \forall_{i\in [t]}\quad \left|\atil_i - \qtil_i(
    D)\right | \leq \gamma,
    \quad \sum_{x\in \cX} D(x) = 1
\end{align}
%

The Lagrangian of \eqref{appex:eq:opt} is :
\begin{align*}
\notag
    \cL &= \diver{D \parallel U} 
    + \sum_{i=1}^t \lambda_i^{+} \pp{\atil_i - \qtil_i(D) - \gamma} 
    + \sum_{i=1}^t \lambda_i^{-}\pp{ \qtil_i(D)- \atil_i - \gamma} 
    + \mu\pp{\sum_{x } D(x) - 1} 
\end{align*}
Let $\lambda \in \mathbb{R}^{t}$ be a vector with, $\lambda_i = \lambda_i^{-} - \lambda_i^{+} $. Then
\begin{align}
    \label{appex:eq:convexDual}
\cL     = \diver{D \parallel U} 
    + \sum_{i=1}^t \lambda_i \qtil_i(D) 
    - \sum_{i=1}^t \lambda_i \atil_i 
    -\gamma \sum_{i=1}^t\pp{\lambda_i^{+} + \lambda_i^{-}}
    + \mu\pp{\sum_{x } D(x) - 1}
\end{align}
where $\|\lambda\|_1 = \sum_{i=1}^t\pp{\lambda_i^{+} + \lambda_i^{-}}$. 
Taking the derivative with respect to $D(x)$ and setting to zero, we get:
\begin{align*}
0 = \frac{\partial\cL }{\partial D(x)} 
&= 
\log\pp{\frac{D(x)}{U(x)}} + 1
 + \sum_{i=1}^t \lambda_i \qtil_i(x)
 + \mu 
\end{align*}

Solving for $D(x)$, we get
\begin{align*}
    D(x) = U(x) \exp\pp{-\sum_{i=1}^t \lambda_i \qtil_i(x) - \mu -1}
\end{align*}

The slack variable $\mu$ must be selected to satisfy the constraint that $\sum_{x\in\cX} D(x)=1$. Therefore have that the solution to \eqref{appex:eq:opt} is a distribution parameterized by the parameter $\lambda$, such that for any $x\in\cX$ we have 
\begin{align*}
  D(x) = \frac{U(x)}{Z} \exp\pp{-\sum_{i=1}^t \lambda_i \qtil_i(x) } 
\end{align*}
where $Z = \sum_{x\in \cX} U(x)\exp\pp{-\sum_{i=1}^t \lambda_i \qtil_i(x) } $.
Plugging into \eqref{appex:eq:convexDual}, we get
\begin{align*}
\cL &= \sum_{x\in \cX} D(x)\log\pp{\frac{D(x)}{U(x)}} 
        + \sum_{i=1}^t \lambda_i \qtil_i(D) 
        - \sum_{i=1}^t \lambda_i \atil_i 
        -\gamma \|\lambda\|_1
       \\ 
       &= \sum_{x\in\cX} D(x) \pp{-\sum_{i=1}^t \lambda_i \qtil_i(x)} 
       - \log(Z)
       +\sum_{i=1}^t \lambda_i \qtil_i(D)
       -\sum_{i=1}^t \lambda_i \atil_i
        -\gamma \|\lambda\|_1
       \\
       &= - \sum_{i=1}^t \lambda_i \qtil_i(D)
       - \log(Z)
            +\sum_{i=1}^t \lambda_i \qtil_i(D)
            -\sum_{i=1}^t \lambda_i \atil_i
             -\gamma \|\lambda\|_1
       \\
       &=  
       - \log(Z)
       + \sum_{i=1}^t \lambda_i\atil_i - \gamma \|\lambda\|_1  \\
        &=  
       - \log\pp{\frac{Z}{\exp\pp{\sum_{i=1}^t \lambda_i\atil_i}}} - \gamma \|\lambda\|_1
       %
\end{align*}

Substituting in for $Z$, we get:

\begin{align*}
\cL &= 
- \log\pp{
\frac{
\sum_{x\in \cX} \exp\pp{\sum_{i=1}^t \lambda_i \qtil_i(x) }
}
{\exp\pp{\sum_{i=1}^t \lambda_i \atil_i }}} - \gamma \|\lambda\|_1 \\
&=
-\log\pp{
\sum_{x\in \cX} \exp\pp{\sum_{i=1}^t \lambda_i \pp{\qtil_i(x) - \atil_i + \gamma}}
}
- \gamma \|\lambda\|_1
\end{align*}

Finally, we have that the dual problem of \eqref{appex:eq:opt} finds a vector $\lambda=(\lambda_1,\ldots, \lambda_t)$ that maximizes $\cL$.  We can write the dual problem as a minimization problem:
\begin{align*}
\cL(\lambda) = 
\min_{\lambda}
\log\pp{
\sum_{x\in \cX} \exp\pp{\sum_{i=1}^t \lambda_i \pp{\qtil_i(x) - \atil_i}}
}
+ \gamma \|\lambda\|_1
\end{align*}




%% file: docs/appendix/pep_iter_oracle.tex
\newcommand{\lambdap}{\lambda^{(+)}}
\newcommand{\lambdam}{\lambda^{(-)}}
\subsection{PEP optimization using iterative projection}

In this section we derive the update rule in algorithm \ref{alg:pep}. Recall that the ultimate goal is to solve \eqref{appex:eq:opt}. Before we describe the algorithm, we remark that it is possible the constraints in problem \ref{appex:eq:opt} cannot be satisfied due to the noise we add to the measurements $\atil_1,\ldots \atil_t$. In principle, $\gamma$ can be chosen to be a high-probability upper bound on the noise, which can be calculated through standard concentration bounds on Gaussian noise. In that case, every constraint $|\atil_i - \qtil_i(D)|\leq \gamma$ can be satisfied and the optimization problem is well defined. However, we note that our algorithm is well defined for every choice of $\gamma\geq 0$. For example, in our experiments we have $\gamma=0$, and we obtain good empirical results that outperform MWEM. In this section we assume that $\gamma=0$.

To explain how algorithm \ref{alg:pep} converges, we cite an established convergence analysis of adaboost from
\cite[chapter 7]{schapire2013boosting} . Similar to adaboost, Algorithm 2 is running iterative projection, where on each iteration, it projects the distribution to satisfy a single constraint. As shown in \cite[chapter 7]{schapire2013boosting}, this iterative algorithm converges to a solution that satisfies all the constraints. The PEP algorithm can be seen as an adaptation of the adaboost algorithm to the setting of query release. Therefore, to solve \eqref{appex:eq:opt}, we use an iterative projection algorithm that on each round selects an unsatisfied constraint and moves the distribution by the smallest possible distance to satisfy it. 

Let $\Qtil_{1:t}$ and $\Atil_{1:t}$ be the set of queries and noisy measurements obtained using the private selection mechanism. Let $K$ be the number of iterations during the optimization and let $D_{t,0}, \ldots, D_{t,K}$ be the sequence of projections during the $K$ iteration of optimization. The goal is that $D_{t,K}$ matches all the constraints defined by $\Qtil_{1:t}$, $\Atil_{1:t}$. Our initial distribution is the uniform distribution $D_{t,0} = U$. Then on round $k \in [K]$, the algorithm selects an index $i_k\in[t]$ such that the $i_k$-th constraint has high error on the current distribution $D_{t,k-1}$. Then the algorithm projects the distribution such that the $i_k$-th constraint is satisfied and the distance to $D_{t,k-1}$ is minimized. Thus, the objective for iteration $k$ is:
\begin{align}\label{eq:privoptIP}
    \text{minimize:}\quad &\diver{D\parallel D_{t,k-1}} 
    & \quad \text{subject to: }\quad & \quad  \atil_{i_k} = \qtil_{i_k}(D), 
    \quad
    \sum_{x\in \cX} D(x) = 1
\end{align}

\input{docs/algos/pep}

 Then the Lagrangian of objective \eqref{eq:privoptIP} is:
\begin{align}\label{eq:lagrangian}
    \cL(D, \lambda) = \diver{D\parallel D_{k-1}} 
    + \lambda \pp{ \qtil_{i_k}(D) - \atil_{i_k}  }
    + \mu\left(\sum_{x\in\cX}D(x) - 1\right) 
\end{align}
Taking the partial derivative with respect to $D(x)$, we have 
\begin{align}\label{eq:dellagrangian}
    \frac{\partial \cL(D, \lambda)}{\partial D(x)} =
    \ln\pp{\frac{D(x)}{D_{k-1}(x)}} + 1 
    +  \lambda  \qtil_{i_k}(x) + \mu =0
\end{align}
Solving \eqref{eq:dellagrangian} for $D(x)$, we get
\begin{align}\label{eq:updatestep}
    D(x) = D_{k-1}(x) \exp\pp{-\lambda \qtil_{i_k}(x) - 1 - \mu} 
    = \tfrac{D_{k-1}(x)}{Z} \exp\pp{-\lambda \qtil_{i_k}(x)} 
\end{align}
where $\mu$ is chosen to satisfy the constraint $\sum_{x\in\cX}D(x) = 1$ and $Z$ is a regularization factor.
Plugging \eqref{eq:updatestep} into \eqref{eq:lagrangian}, we get:
\begin{align*}
    \cL(D, \lambda) &= \diver{D \parallel D_{k-1}} 
    + \lambda \pp{ \qtil_{i_k}(D) - \atil_{i_k}  } \\
    &= \sum_{x} D(x) \log\pp{ \frac{D(x)}{D_{k-1}(x)}}
        +\lambda \left( \qtil_{i_k}(D) - \atil_{i_k}  \right) \\ 
     &= \sum_{x} D(x) \log\pp{\frac{1}{Z} \frac{D_{k-1}(x)e^{-\lambda q_{i_k}(x)}}{D_{k-1}(x)}} 
        +\lambda \left( \qtil_{i_k}(D) - \atil_{i_k}  \right) \tag{\ref{eq:updatestep}}\\ 
    %
    %
    %
    &= -\lambda\sum_{x} D(x)\qtil_{i_k}(x)
    - \log(Z)
        +\lambda \left( \qtil_{i_k}(D) - \atil_{i_k}  \right) \\ 
    &= -\lambda \qtil_{i_k}(D)
    - \log(Z)
        +\lambda \left( 
        \qtil_{i_k}(D) - \atil_{i_k} \right) \\ 
    &=  -\log(Z) - \lambda\atil_{i_k}   \\ 
    &=  -\log\pp{\sum_{x\in\cX}D_{k-1}(x)\exp\pp{-\lambda \qtil_{i_k}(x)} } - \lambda \atil_{i_k}  \\ 
\end{align*}

The next step is to find the optimal value of $\lambda$. Therefore we calculate the derivative of $\cL(D, \lambda) $ with respect to $\lambda$:
\begin{align*}
    \frac{\partial \cL(D, \lambda)}{\partial \lambda} 
    &= \frac{e^{-\lambda} \qtil_{i_k}(D_{k-1})}{\sum_{x\in\cX } D_{k-1}(x) \exp\pp{-\lambda \qtil_{i_k}(x)} } - \atil_{i_k}\\
    &= \frac{e^{-\lambda} \qtil_{i_k}(D_{k-1})}{e^{-\lambda} \qtil_{i_k}(D_{k-1}) + (1-\qtil_{i_k}(D_k)) }- \atil_{i_k}
\end{align*}

Setting $\frac{\partial \cL(D, \lambda)}{\partial \lambda}  = 0$, we can solve for $\lambda$.
\begin{align*}
\frac{e^{-\lambda} \qtil_{i_k}(D_{k-1})}{e^{-\lambda} \qtil_{i_k}(D_{k-1}) + (1-\qtil_{i_k}(D_k)) } &=  \atil_{i_k} \\ 
e^{-\lambda} \qtil_{i_k}(D_{k-1}) &=  \atil_{i_k}\pp{e^{-\lambda} \qtil_{i_k}(D_{k-1}) + (1-\qtil_{i_k}(D_k)) } \\ 
e^{-\lambda} \qtil_{i_k}(D_{k-1}) - e^{-\lambda} \atil_{i_k}\qtil_{i_k}(D_{k-1})&=  \atil_{i_k}\pp{ 1-\qtil_{i_k}(D_k) } \\ 
e^{-\lambda} \qtil_{i_k}(D_{k-1})\pp{1 - \atil_{i_k}} 
&=  \atil_{i_k}\pp{ 1-\qtil_{i_k}(D_k) } \\ 
\end{align*}
Finally we obtain
\begin{align*}
    -\lambda = \ln\pp{\frac{\atil_{i_k} \pp{1- \qtil_{i_k}(D_{k-1}) } }{ (1-\atil_{i_k})\qtil_{i_k}(D_{k-1}) }}
\end{align*}

%% file: docs/algos/pep.tex
\begin{algorithm}[H]
\SetAlgoLined
\textbf{Input:}
Error tolerance $\gamma$, linear queries $\Qtil_{1:T} = \{\qtil_1, \ldots, \qtil_T\}$,  and noisy measurements $\Atil_{1:T} = \{\atil_1, \ldots, \atil_T \}$. \\
\textbf{Objective:} Minimize $ \diver{D \parallel U} $  such that  $\forall_{i \in [T]} \quad \left| \qtil_i(D) -  \atil_i\right| \leq \gamma$. \\
Initialize $D_0$ to be the uniform distribution over $\cX$, and $t \leftarrow 0 $. \\ 
 \While{$\max_{i\in[T]} \left| \ahat_i - \qtil_i(D_t) \right| > \gamma$ }{
  \textbf{Choose:} $i \in [T]$  with 
  $i \leftarrow \argmax_{j\in[T]}  \left|\atil_j - \qtil_j(D_t) \right|$. \\
  \textbf{Update:} For all $x\in\cX$, set  $D_{t+1}(x) \leftarrow D_{t}(x) e^{-\lambda_t \qtil_i(x)}$, 
  where 
  $-\lambda_t = \ln\left( \frac{\atil_i  (1-\qtil_i(D_t)) }{(1-\atil_i) \qtil_i(D_t)} \right)$. \\ 
  $t \leftarrow t + 1$ \\ 
 }
 \textbf{Output:}  $D_T$
 \caption{Exponential Weights Projection}
 \label{alg:pep}
\end{algorithm}

%% file: docs/appendix/experiments.tex
\section{Additional empirical evaluation}


\subsection{Experimental details}\label{appx:experiments}

We present hyperparameters used for methods across all experiments in Tables \ref{tab:hyperparameters_pep}, \ref{tab:hyperparameters_gem}, \ref{tab:hyperparameters_pep_pub}, \ref{tab:hyperparameters_gem_pub}, and \ref{tab:hyperparameters_baselines}. To limit the runtime of \pep and \peppub, we add the hyperparameter, $T_{max}$, which controls the maximum number of update steps taken at each round $t$. Our implementations of \mwem, \dq, and \pmwpub are adapted from \url{https://github.com/terranceliu/pmw-pub}. We implement \rap and \rapsoftmax ourselves using PyTorch since the code for \rap. All experiments are run using a desktop computer with an Intel® Core™ i5-4690K processor and NVIDIA GeForce GTX 1080 Ti graphics card.

We obtain the ADULT and ACS datasets by following the instructions outlined in \url{https://github.com/terranceliu/pmw-pub}. Our version of ADULT used to train \gempubred (Figure \ref{fig:max_error_pub}b and \ref{fig:all_errors_adult_pub}) uses the following attributes: sex, race, relationship, marital-status, occupation, education-num, age.

\input{tables/hyperparameters}

\subsection{Main experiments with additional metrics}

\input{figures/all_errors/acs_adult}

\input{figures/all_errors/acs_pub}

\input{figures/all_errors/adult_pub}

In Figures \ref{fig:all_errors_acs_adult}, \ref{fig:all_errors_acs_pub}, and \ref{fig:all_errors_adult_pub}, we present the same results for the same experiments described in Section \ref{sec:results} (Figures \ref{fig:max_error_acs_adult} and \ref{fig:max_error_pub}), adding plots for mean error and root mean squared error (RMSE). For our experiments on ACS PA-18 with public data, we add results using 2018 data for Ohio (ACS OH-18), which we note also low \textit{best-mixture-error}. Generally, the relative performance between the methods for these other two metrics is the same as for max error.

In addition, in Figure \ref{fig:all_errors_acs_pub}, we present results for \peppub, a version of \pep similar to \pmwpub that is adapted to leverage public data (and consequently can be applied to high dimensional settings). We briefly describe the details below.

\peppub. Like in \citet{liu2021leveraging}, we extend \pep by making two changes: (1) we maintain a distribution over the public data domain and (2) we initialize the approximating distribution to that of the public dataset. Therefore like \pmwpub, \peppub also restricts $\cD$ to distributions over the public data domain and initializes $D_0$ to be the public data distribution.

We note that \peppub performs similarly to \pmwpub, making it unable to perform well when using ACS CA-18 as a public dataset (for experiments on ACS PA-18). Similarly, it cannot be feasibly run for the ADULT dataset when the public dataset is missing a significant number of attributes. 


\subsection{Comparisons against \rap}\label{appx:rap_compare}

\input{figures/all_errors/rap_rapsoftmax}

In Figure \ref{fig:all_errors_rap_rapsoftmax}, we show failures cases for \rap. Again, we see that \rapsoftmax outperforms \rap in every setting. However, we observe that aside from ADULT (reduced), \rap performs extremely poorly across all privacy budgets. 

To account for this observation, we hypothesize that by projecting each measurement to \citet{aydore2021differentially}'s proposed continuous relaxation of the synthetic dataset domain, \rap produces a synthetic dataset that is inconsistent with the semantics of an actual dataset. Such inconsistencies make it more difficult for the algorithm to do well without seeing the majority of high error queries.

Consider this simple example comparing \gem and \rap. Suppose we have some binary attribute $A \in \curlybrack{0, 1}$ and we have $P(A=0) = 0.2$ and $P(A=1) = 0.8$. For simplicity, suppose that the initial answers at $t=0$ for both algorithms is $0$ for the queries $q_{A=0}$ and $q_{A=1}$. Assume at $t=1$ that the privacy budget is large enough such that both algorithms select the max error query $q_{A=1}$ (error of $0.8$), which gives us an error or $0.8$. After a single iteration, both algorithms can reduce the error of this query to $0$. In \rap, the max error then is $0.2$ (for the next largest error query $q_{A=0}$). However for \gem to output the correct answer for $q_{A=1}$, it must learn a distribution (due to the softmax activation function) such that $P(A=1) = 0.8$, which naturally forces $P(A=0) = 0.2$. In this way, \gem can reduce the errors of both queries in one step, giving it an advantage over \rap. 

In general, algorithms within the \adaptive framework have this advantage in that the answers it provides must be consistent with the data domain. For example, if again we consider the two queries for attribute $A$, a simple method like the Gaussian or Laplace mechanism has a nonzero probability of outputting noisy answers for $q_{A=0}$ and $q_{A=1}$ such that $P(A=0) + P(A=1) \ne 1$. This outcome however will never occur in \adaptive.

Therefore, we hypothesize that \rap tends to do poorly as you increase the number of high error queries because the algorithm needs to select each high error query to obtain low error. Synthetic data generation algorithms can more efficiently make use of selected query measurements because their answers to all possible queries must be consistent. Referring to the above example again, there may exist two high error queries $q_{A=0}$ and $q_{A=1}$, but only one needs to be sampled to reduce the errors of both.

We refer readers to Appendix \ref{appx:adult_loans}, where we use the above discussion to account for how the way in which the continuous attributes in ADULT are preprocessed can impact the effectiveness of \rap.


\subsection{Marginal trick}\label{appx:experiments_marginal_trick}

While this work follows the literature in which methods iteratively sample sensitivity $1$ queries, we note that the marginal trick approach (Appendix \ref{appx:marginal_trick}) can be applied to all iterative algorithms under \adaptive. To demonstrate this marginal trick's effectiveness, we show in Figure \ref{fig:all_errors_gem_marginal} how the performance of \gem improves across max, mean, and root mean squared error by replacing the \textit{Update} and \textit{Measure} steps in \adaptive. 

In this experiment, given that the number of measurements taken is far greater when using the marginal trick, we increased $T_{max}$ for \gem from $100$ to $10000$ and changed the loss function from $\ell_1$-loss to $\ell_2$-loss. Additional hyperparameters used can be found in Table \ref{tab:hyperparameters_gem_marginal}. Note that we reduced the model size for $G$ simply to speed up runtime. Overall, we admit that leveraging this trick was not our focus, and so we leave designing \gem (and other iterative methods) to fully take advantage of the marginal trick to future work. 

\input{tables/hyperparameters/hyperparameters_gem_marginal}

\input{figures/all_errors/gem_marginal}


\subsection{Discussion of \hdmm}\label{appx:hdmm}

\hdmm \citep{mckenna2018optimizing} is an algorithm designed to directly answer a set of workloads, rather than some arbitrary set of queries. In particular, \hdmm optimizes some strategy matrix to represent each workload of queries that in theory, facilitates an accurate reconstruction of the workload answers while decreasing the sensitivity of the privacy mechanisms itself. In their experiments, \citet{mckenna2018optimizing} show strong results w.r.t. RMSE, and the U.S. Census Bureau itself has incorporated aspects of the algorithm into its own releases \citep{Kifer19}.

We originally planned to run \hdmm as a baseline for our algorithms in the standard setting, but after discussing with the original authors, we learned that currently, the available code for \hdmm makes running the algorithm difficult for the ACS and ADULT datasets. There is no way to solve the least square problem described in the paper for domain sizes larger than $10^9$, and while the authors admit that \hdmm could possibly be modified to use local least squares for general workloads (outside of those defined in their codebase), this work is not expected to be completed in the near future.

We also considered running \hdmmpgm \citep{mckenna2019graphical}, which replaces the least squares estimation problem a graphical model estimation algorithm. Specifically, using (differentially private) measurements to some set of input queries, \hdmmpgm infers answers for any workload of queries. However, the memory requirements of the algorithm scale exponentially with dimension of the maximal clique of the measurements, prompting users to carefully select measurements that help build a useful junction tree that is not too dense. Therefore, the choice of measurements and cliques can be seen as hyperparameters for \hdmmpgm, but as the authors pointed out to us, how such measurements should be selected is an open problem that hasn't been solved yet. In general, cliques should be selected to capture correlated attributes without making the size of the graphical model intractable. However, we were unsuccessful in finding a set of measurements that achieved sensible results (possibly due to the large number of workloads our experiments are designed to answer) and decided stop pursuing this endeavor due to the heavy computational resources required to run \hdmmpgm. We leave finding a proper set of measurements for ADULT and ACS PA-18 as an open problem.

Given such limitations, we evaluate \hdmm with least squares on ACS (reduced) PA-18 and ADULT (reduced) only (Figure \ref{fig:all_errors_hdmm}). We use the implementation found in \url{https://github.com/ryan112358/private-pgm}. We compare to \gem using the marginal trick, which \hdmm also utilizes by default. While \gem outperforms \hdmm, \hdmm seems to be very competitive on low dimensional datasets when the privacy budget is higher. In particular, \hdmm slightly outperforms \gem w.r.t. max error on ACS (reduced) when $\varepsilon=1$. We leave further investigation of \hdmm and \hdmmpgm to future work.

\input{figures/all_errors/hdmm}


\subsection{Effectiveness of optimizing over past queries}

\input{figures/all_errors/mwem_no_opt}

One important part of the adaptive framework is that it encompasses algorithms whose update step uses measurements from past iterations. In Figure \ref{fig:all_errors_mwem_no_opt}, we verify claims from \citet{hardt2010simple} and \citet{liu2021leveraging} that empirically, we can significantly improve over the performance of \mwem when incorporating past measurements.


\subsection{Evaluating on ADULT* and LOANS}\label{appx:adult_loans}

\input{figures/all_errors/adult_loans}

In Figure \ref{fig:all_errors_adult_loans}, we reproduce the experiments on the ADULT (which we denote as ADULT*) and LOANS datasets presented in \citet{aydore2021differentially}. Like \citet{aydore2021differentially}, we obtain the datasets from \url{https://github.com/giusevtr/fem}. Empirically, we find that \gem outperforms all baseline methods. In addition, while \rap performs reasonably well, we observe that by confining $\cD$ to $\curlybrack{\sigma(M) | M \in \mathbb{R}^{n' \times d}}$ with the softmax function, \rapsoftmax performs better across all privacy budgets.

To account for why \rap performs reasonably well with respect to max error on ADULT* and LOANs but very poorly on ADULT and ACS, we refer back to our discussion about the issues of \rap presented in Appendix \ref{appx:rap_compare} in which argue that by outputting synthetic data that is inconsistent with any real dataset, \rap performs poorly when there are many higher error queries. ADULT* and LOANs are preprocessed in a way such that continous attributes are converted into categorical (technically ordinal) attributes, where a separate categorical value is created for each unique value that the continuous attribute takes on in the dataset (up to a maximum of $100$ unique values). When processed in this way, $k$-way marginal query answers are sparser, even when $k$ is relatively small ($ \le 5$). However, \citet{liu2021leveraging} preprocess continuous variables in the ADULT and ACS dataset by constructing bins, resulting in higher error queries.

For example, suppose in an unprocessed dataset (with $n$ rows), you have $3$ rows where an attribute (such as income) takes on the values $16,587$, $15,984$, and $18,200$. Next, suppose there exists datasets A and B, where dataset A maps each unique value to its own category, while dataset B constructs a bin for values between $15,000$ and $20,000$. Then considering all $1$-way marginal queries involving this attribute, dataset A would have $3$ different queries, each with answer $\frac{1}{N}$. Dataset B however would only have a single query whose answer is $\frac{3}{N}$. Whether a dataset should be preprocessed as dataset A or dataset B depends on the problem setting.\footnote{We would argue that in many cases, dataset B makes more sense since it is more likely for someone to ask---"How many people make between $15,000$ and $20,000$ dollars?"---rather than---"How many people make $15,984$ dollars?".} However, this (somewhat contrived) example demonstrates how dataset B would have more queries with high value answers (and therefore more queries with high initial errors, assuming that the algorithms in question initially outputs answers that are uniform/close to $0$). 

In our experiments with $3$-way marginal queries, ADULT (where workload is $286$) and ADULT* (where the workload is $64$) have roughly the same number queries ($334,128$ vs. $458,996$ respectively). However, ADULT has $487$ queries with answers above $0.1$ while ADULT* only has $71$. Looking up the number of queries with answers above $0.2$, we count $181$ for ADULT and only $28$ for ADULT*. Therefore, experiments on ADULT* have fewer queries that \rap needs to optimize over to achieve low max error, which we argue accounts for the differences in performance on the two datasets.

Finally, we note that in Figure \ref{fig:all_errors_rap_rapsoftmax}, \rap has relatively high mean error and RMSE. We hypothesize that again, because only the queries selected on each round are optimized and all other query answers need not be consistent with the optimized ones, \rap will not perform well on any metric that is evaluated over all queries (since due to privacy budget constraints, most queries/measurements are never seen by the algorithm). We leave further investigation on how \rap operates in different settings to future work.

%% file: tables/hyperparameters.tex
\input{tables/hyperparameters/hyperparameters_pep}

\input{tables/hyperparameters/hyperparameters_gem}

\input{tables/hyperparameters/hyperparameters_pep_pub}

\input{tables/hyperparameters/hyperparameters_gem_pub}

\input{tables/hyperparameters/hyperparameters_baselines}

%% file: tables/hyperparameters/hyperparameters_pep.tex
\begin{table}[!h]
\centering
\caption{\pep hyperparameters}
\label{tab:hyperparameters_pep}
\begin{tabular}{l l c}
    \toprule
    Dataset & Parameter & Values \\
    \midrule
    \multirow{1}{*}{All}
    & {$T_{max}$} & $25$ \\
    \midrule
    \multirow{2}{*}{ACS (red.)} 
    & \multirow{2}{*}{$T$} 
    & $20$, $30$, $40$, $50$, $75$ \\
    & & $100$, $125$, $150$, $175$, $200$ \\
    \midrule
    \multirow{2}{*}{ADULT (red.)} 
    & \multirow{2}{*}{$T$} 
    & $20$, $30$, $40$, $50$, $75$ \\
    & & $100$, $125$, $150$, $175$, $200$ \\
    \bottomrule
\end{tabular}
\end{table}

%% file: tables/hyperparameters/hyperparameters_gem.tex
\begin{table}[!h]
\centering
\caption{\gem hyperparameters}
\label{tab:hyperparameters_gem}
\begin{tabular}{l l c}
    \toprule
    Dataset & Parameter & Values \\
    \midrule
    \multirow{4}{*}{All} 
    & hidden layer sizes & $(512, 1024, 1024)$ \\
    & learning rate & $0.0001$ \\
    & $B$ & $1000$ \\
    & $\alpha$ & $0.67$ \\
    & {$T_{max}$} & $100$ \\
    \midrule
    \multirow{2}{*}{ACS} 
    & \multirow{2}{*}{$T$} 
    & $100$, $150$, $200$, $250$, $300$, \\
    & & $400$, $500$, $750$, $1000$ \\
    \midrule
    \multirow{2}{*}{ACS (red.)} 
    & \multirow{2}{*}{$T$} 
    & $50$, $75$, $100$, $125$, $150$, \\
    & & $200$, $250$, $300$ \\
    \midrule
    ADULT, ADULT (red.), & \multirow{3}{*}{$T$} 
    & $30$, $40$, $50$, $60$, $70$, \\
    ADULT (orig), &
    & $80$, $90$, $100$, $125$, $150$, \\
    LOANS & & $175$, $200$ \\
    \bottomrule
\end{tabular}
\end{table}

%% file: tables/hyperparameters/hyperparameters_pep_pub.tex
\begin{table}[!h]
\centering
\caption{\peppub hyperparameters}
\label{tab:hyperparameters_pep_pub}
\begin{tabular}{l l c}
    \toprule
    Dataset & Parameter & Values \\
    \midrule
    \multirow{1}{*}{All}
    & {$T_{max}$} & $25$ \\
    \midrule
    \multirow{2}{*}{ACS} 
    & \multirow{2}{*}{$T$} 
    & $20$, $40$, $60$, $80$, $100$ \\
    & & $120$, $140$, $160$, $180$, $200$ \\
    \bottomrule
\end{tabular}
\end{table}

%% file: tables/hyperparameters/hyperparameters_gem_pub.tex
\begin{table}[!h]
\centering
\caption{\gempub hyperparameters}
\label{tab:hyperparameters_gem_pub}
\begin{tabular}{l l c}
    \toprule
    Dataset & Parameter & Values \\
    \midrule
    \multirow{4}{*}{All} 
    & hidden layer sizes & $(512, 1024, 1024)$ \\
    & learning rate & $0.0001$ \\
    & $B$ & $1000$ \\
    & $\alpha$ & $0.67$ \\
    & {$T_{max}$} & $100$ \\
    \midrule
    \multirow{2}{*}{ACS} 
    & \multirow{2}{*}{$T$} 
    & $30$, $40$, $50$, $75$, $100$, \\
    & & $150$, $200$, $300$, $400$, $500$ \\
    \midrule
    \multirow{2}{*}{ADULT} 
    & \multirow{2}{*}{$T$}
    & $2$, $3$, $5$, $10$, $20$, $30$, $40$, $50$, \\
    & & $60$ $70$ $80$ $90$ $100$ \\
    \bottomrule
\end{tabular}
\end{table}

%% file: tables/hyperparameters/hyperparameters_baselines.tex
\begin{table}[!h]
\centering
\caption{Baseline hyperparameters}
\label{tab:hyperparameters_baselines}
\begin{tabular}{l l c}
    \toprule
    Method & Parameter & Values \\
    \midrule
    \multirow{4}{*}{\rap} 
    & learning rate & $0.001$ \\
    & $n'$ & $1000$ \\
    & $K$ & $5$, $10$, $25$, $50$, $100$ \\
    & $T$ & $2$, $5$, $10$, $25$, $50$, $75$, $100$ \\
    \midrule
    \multirow{4}{*}{\rapsoftmax} 
    & learning rate & $0.1$ \\
    & $n'$ & $1000$ \\
    & $K$ & $5$, $10$, $25$, $50$, $100$ \\
    & $T$ & $2$, $5$, $10$, $25$, $50$, $75$, $100$ \\
    \midrule
    \multirow{2}{*}{\mwem}
    & \multirow{2}{*}{$T$} 
    & $100$, $150$, $200$, $250$, $300$ \\
    & & $400$, $500$, $750$, $1000$ \\
    \midrule
    \multirow{1}{*}{\mwem (/w past queries)}
    & $T$ & $50$, $75$, $100$, $150$, $200$, $250$, $300$ \\
    \midrule
    \multirow{2}{*}{\dq}
    & $\eta$ & $2$, $3$, $4$, $5$ \\
    & samples & $25$ $50$, $100$, $250$, $500$ \\
    \bottomrule
\end{tabular}
\end{table}

%% file: figures/all_errors/acs_adult.tex
\begin{figure}[!t]
    \centering
    \includegraphics[width=\linewidth]{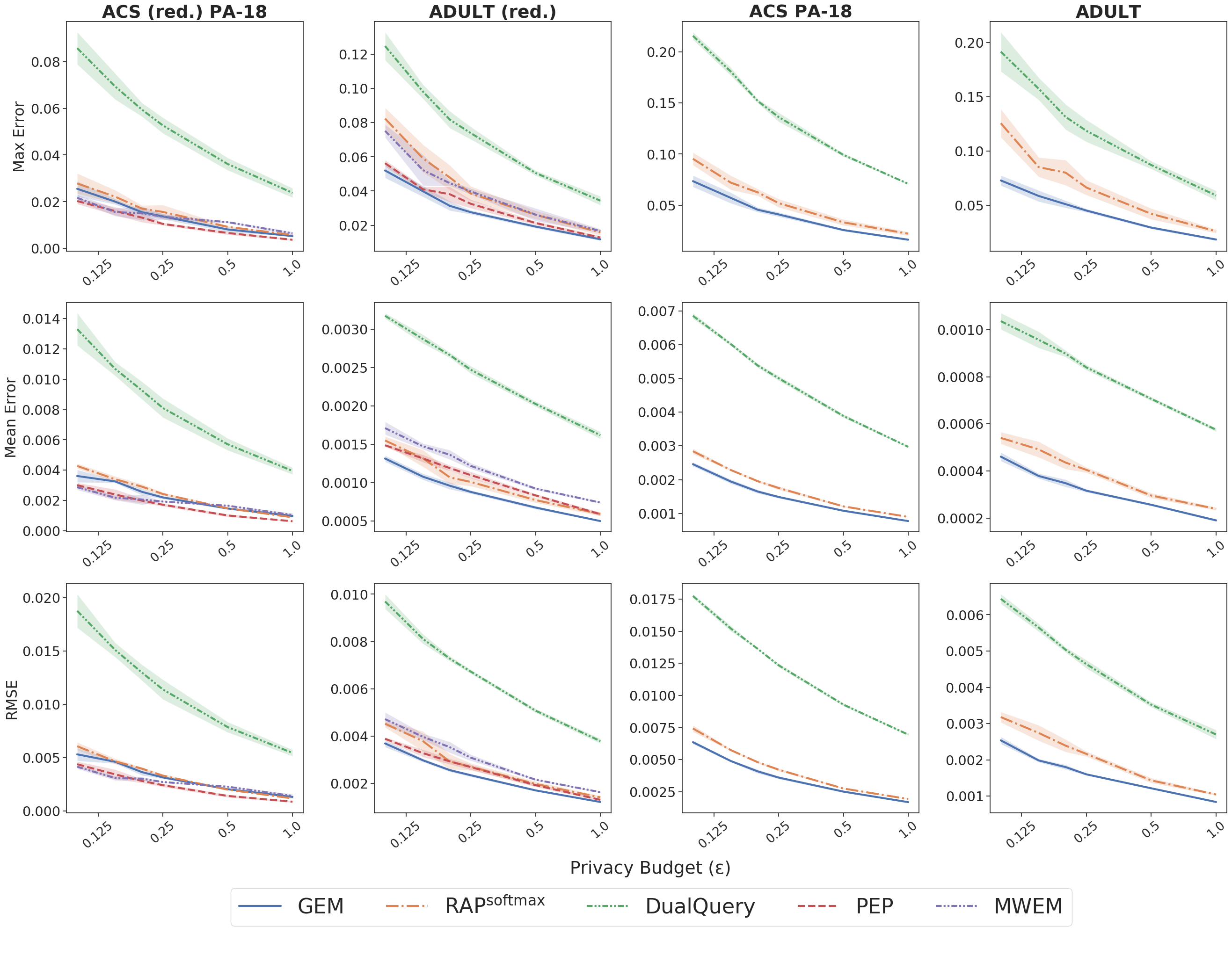}
    \caption{
    Max, mean, and root mean squared errors for $3$-way marginals evaluated on ADULT and ACS PA-18 using privacy budgets $\varepsilon \in \{ 0.1, 0.15, 0.2, 0.25, 0.5, 1 \}$ and $\delta = \frac{1}{n^2}$. The \textit{x-axis} uses a logarithmic scale. We evaluate using the following workload sizes:
    ACS (reduced) PA-18: $455$;
    ADULT (reduced): $35$; 
    ACS PA-18: $4096$; 
    ADULT: $286$.
    Results are averaged over $5$ runs, and error bars represent one standard error.
    }
    \label{fig:all_errors_acs_adult}
\end{figure}

%% file: figures/all_errors/acs_pub.tex
\begin{figure}[!t]
    \centering
    \includegraphics[width=\linewidth]{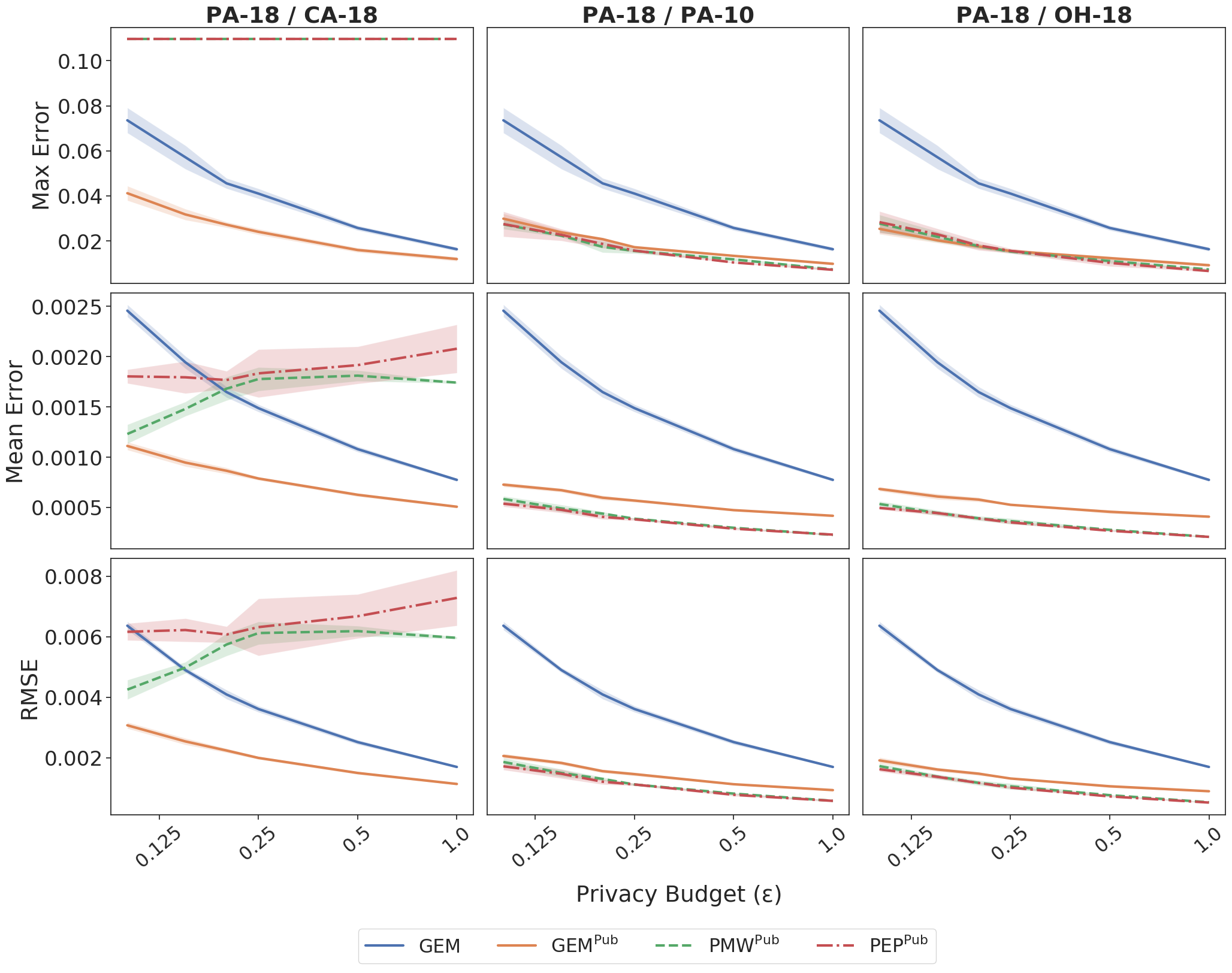}
    \caption{
    Max, mean, and mean squared error for $3$-way marginals on ACS PA-18 (workloads = $4096$) with privacy budgets $\varepsilon \in \{ 0.1, 0.15, 0.2, 0.25, 0.5, 1 \}$ and $\delta = \frac{1}{n^2}$. We evaluate public-data-assisted algorithms with the following public datasets:
    \textbf{Left:} 2018 California (CA-18);
    \textbf{Center:} 2010 Pennsylvania (PA-10);
    \textbf{Right:} 2018 Ohio (PA-10).
    The \textit{x-axis} uses a logarithmic scale. Results are averaged over $5$ runs, and error bars represent one standard error. 
    }
    \label{fig:all_errors_acs_pub}
\end{figure}

%% file: figures/all_errors/adult_pub.tex
\begin{figure}[!t]
    \centering
    \includegraphics[width=\linewidth]{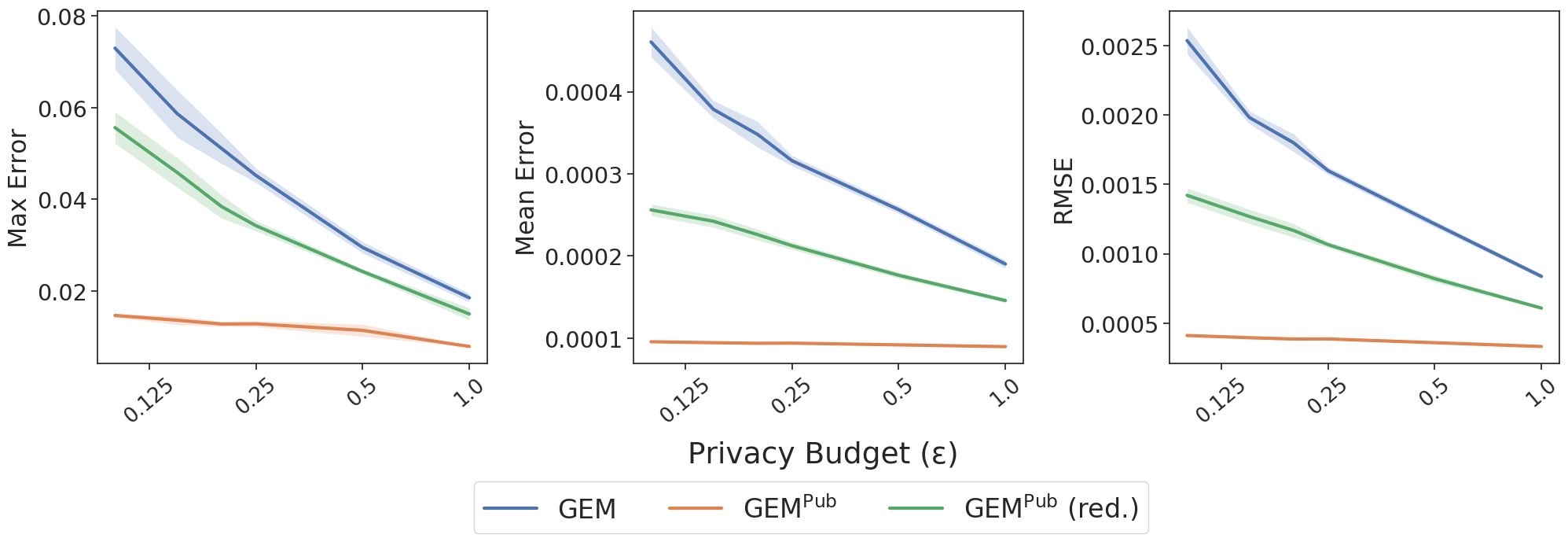}
    \caption{
    Max, mean, and mean squared error for $3$-way marginals on ADULT (workloads = $286$) with privacy budgets $\varepsilon \in \{ 0.1, 0.15, 0.2, 0.25, 0.5, 1 \}$ and $\delta = \frac{1}{n^2}$. We evaluate \gem using both the complete public data (\gempub) and a reduced version that has fewer attributes (\gempubred).
    The \textit{x-axis} uses a logarithmic scale. Results are averaged over $5$ runs, and error bars represent one standard error. 
    }
    \label{fig:all_errors_adult_pub}
\end{figure}

%% file: figures/all_errors/rap_rapsoftmax.tex
\begin{figure}[!t]
    \centering
    \includegraphics[scale=0.3]{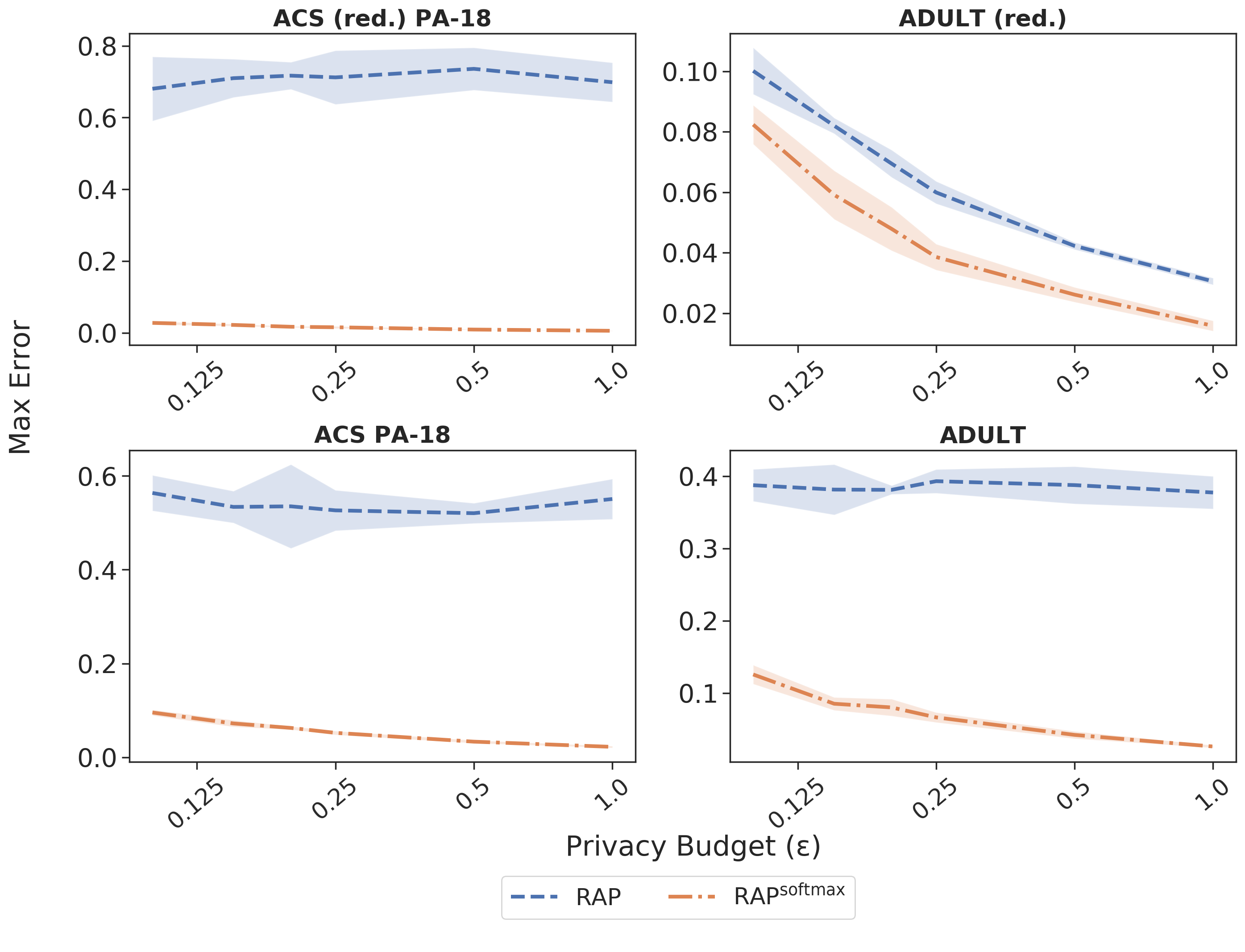}
    \caption{
    Comparison of \rap and \rapsoftmax w.r.t max error for $3$-way marginals evaluated on ADULT and ACS PA-18 using privacy budgets $\varepsilon \in \{ 0.1, 0.15, 0.2, 0.25, 0.5, 1 \}$ and $\delta = \frac{1}{n^2}$. The \textit{x-axis} uses a logarithmic scale. We evaluate using the following workload sizes:
    ACS (reduced) PA-18: $455$;
    ACS PA-18: $4096$; 
    ADULT (reduced): $286$; 
    ADULT: $35$.
    Results are averaged over $5$ runs, and error bars represent one standard error.
    }
    \label{fig:all_errors_rap_rapsoftmax}
\end{figure}

%% file: tables/hyperparameters/hyperparameters_gem_marginal.tex
\begin{table}[!h]
\centering
\caption{\gem (marginal trick) hyperparameters}
\label{tab:hyperparameters_gem_marginal}
\begin{tabular}{l l c}
    \toprule
    Dataset & Parameter & Values \\
    \midrule
    \multirow{4}{*}{All} 
    & hidden layer sizes & $(256, 512)$ \\
    & learning rate & $0.0001$ \\
    & $B$ & $1000$ \\
    & $\alpha$ & $0.5$ \\
    & {$T_{max}$} & $10000$ \\
    \midrule
    \multirow{2}{*}{ACS} 
    & \multirow{2}{*}{$T$} 
    & $50$, $100$, $150$, $200$, \\
    & & $250$, $300$, $400$, $500$ \\
    \midrule
    \multirow{2}{*}{ACS (red.)} 
    & \multirow{2}{*}{$T$} 
    & $50$, $100$, $150$, $200$, \\
    & & $250$, $300$, $450$ \\
    \midrule
    \multirow{2}{*}{ADULT} 
    & \multirow{2}{*}{$T$} 
    & $30$, $40$, $50$, $75$, \\
    & & $100$, $125$, $150$, $200$ \\
    \midrule
    \multirow{2}{*}{ADULT (red.)} 
    & \multirow{2}{*}{$T$} 
    & $5$, $10$, $15$, $20$ \\
    & & $25$, $30$, $35$ \\
    \bottomrule
\end{tabular}
\end{table}

%% file: figures/all_errors/gem_marginal.tex
\begin{figure}[!t]
    \centering
    \includegraphics[width=\linewidth]{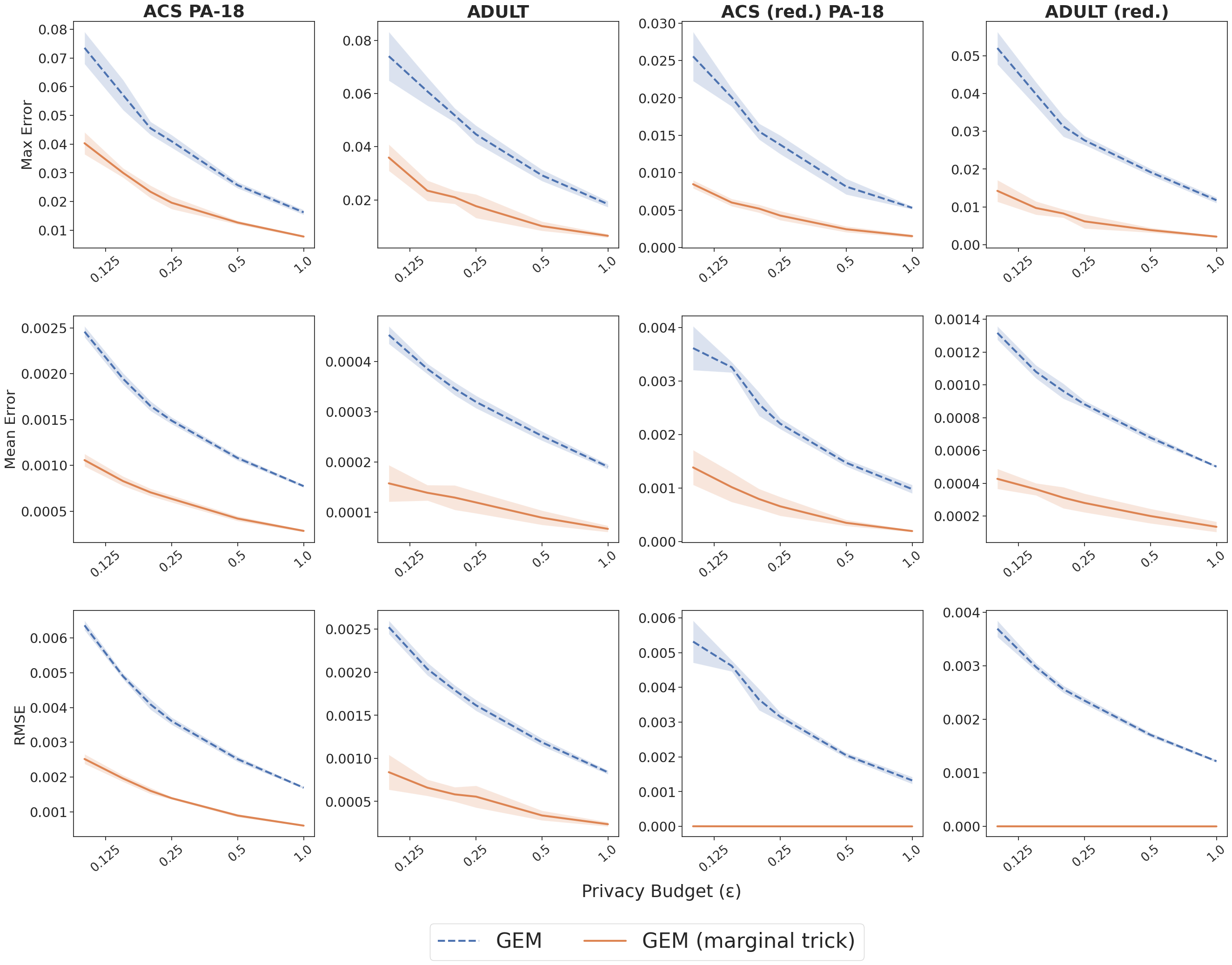}
    \caption{
    Error comparison of \gem with and without the marginal trick , evaluated on $3$-way marginals with privacy budgets $\varepsilon \in \{ 0.1, 0.15, 0.2, 0.25, 0.5, 1 \}$ and $\delta = \frac{1}{n^2}$. The \textit{x-axis} uses a logarithmic scale. Results are averaged over $5$ runs, and error bars represent one standard error.
    }
    \label{fig:all_errors_gem_marginal}
\end{figure}

%% file: figures/all_errors/hdmm.tex
\begin{figure}[!t]
    \centering
    \includegraphics[scale=0.25]{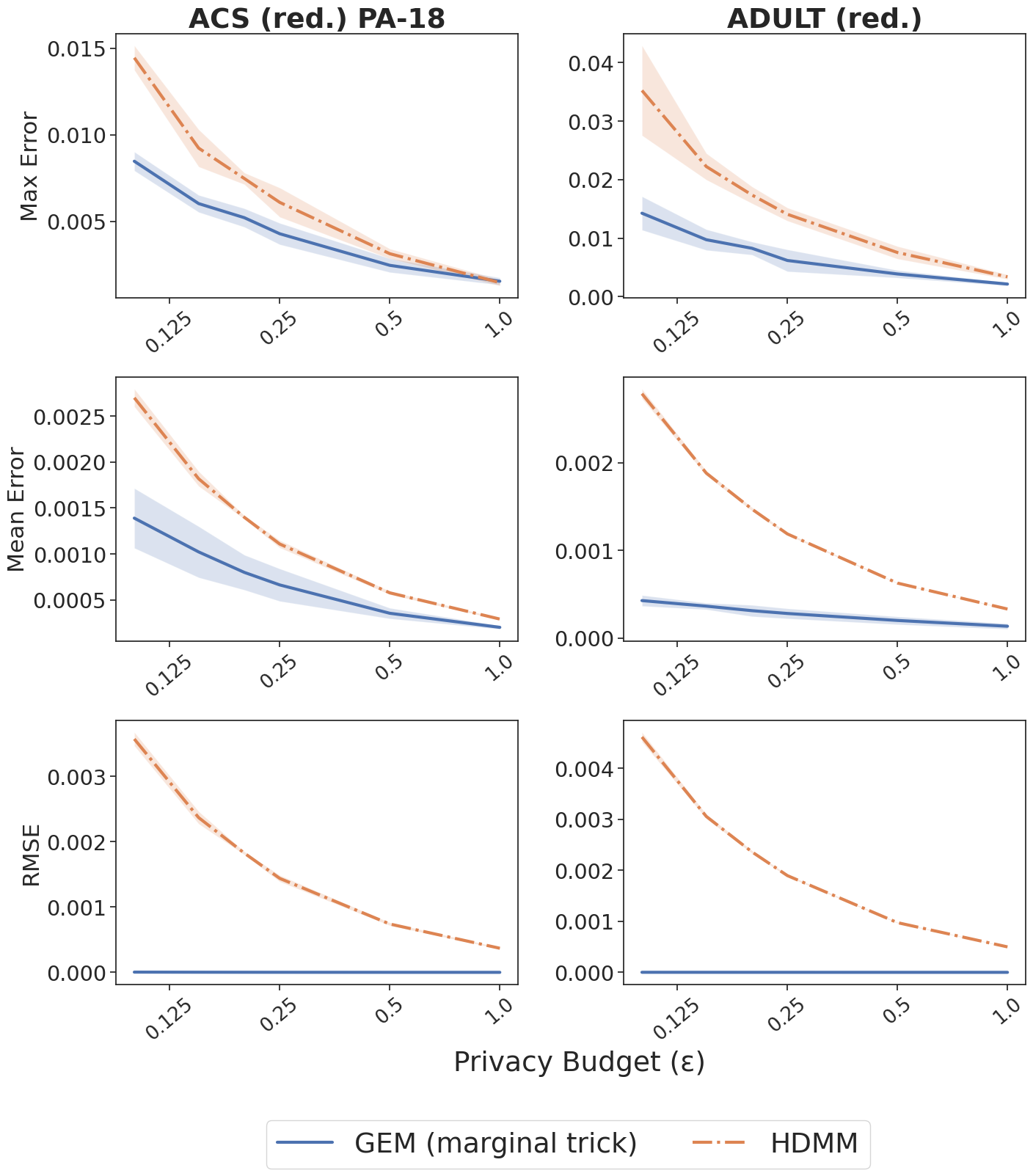}
    \caption{
    Comparison of max, mean, and root mean squared errors against \hdmm on ACS (reduced) PA-18 (workloads=455)  and ADULT (reduced) (workloads=35), evaluated on $3$-way marginals with privacy budgets $\varepsilon \in \{ 0.1, 0.15, 0.2, 0.25, 0.5, 1 \}$ and $\delta = \frac{1}{n^2}$. The \textit{x-axis} uses a logarithmic scale. Results are averaged over $5$ runs, and error bars represent one standard error.
    }
    \label{fig:all_errors_hdmm}
\end{figure}

%% file: figures/all_errors/mwem_no_opt.tex
\begin{figure}[!t]
    \centering
    \includegraphics[width=\linewidth]{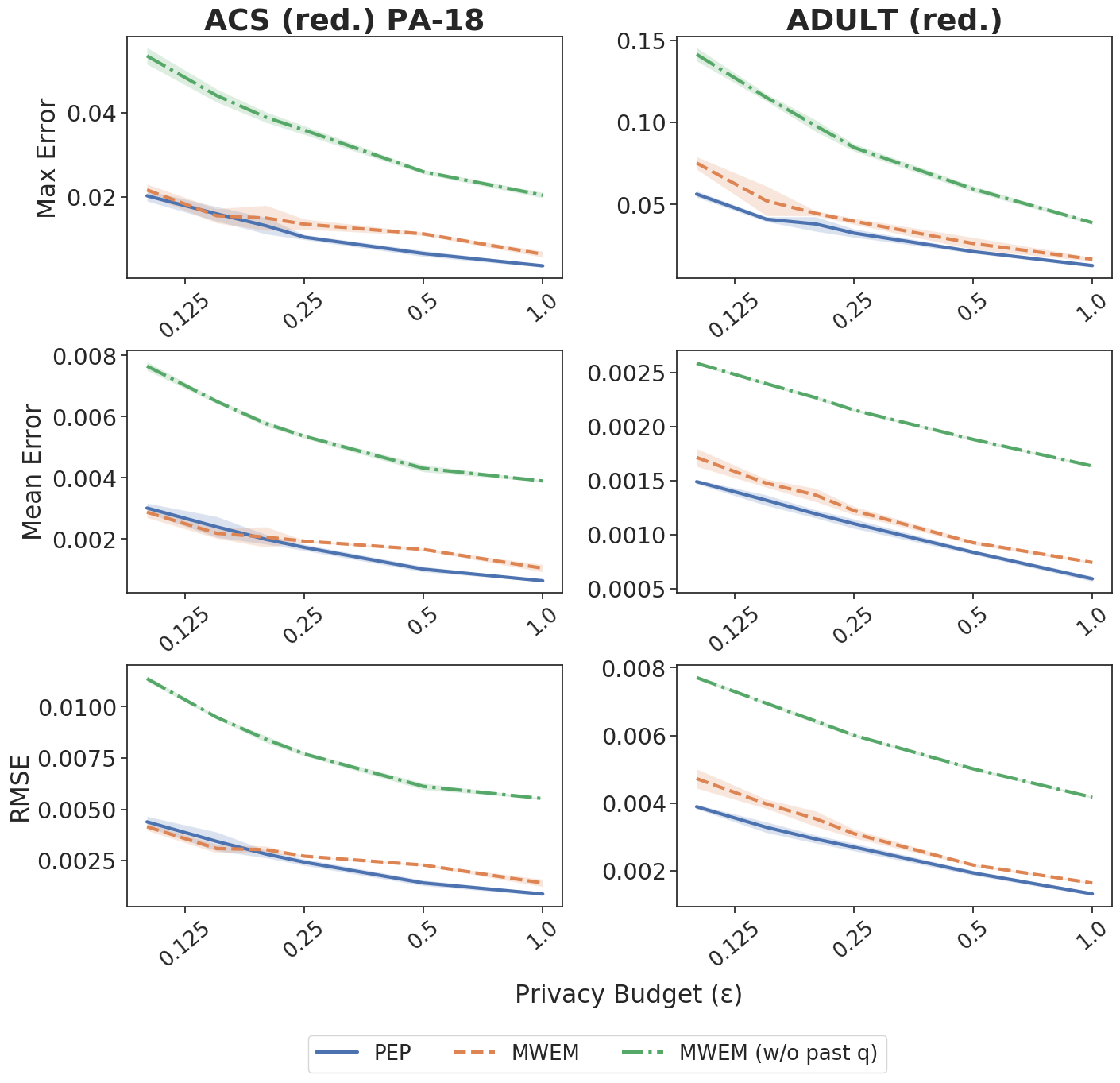}
    \caption{
    Comparison of max, mean, and root mean squared errors against vanilla \mwem that does not use queries sampled during past iterations, evaluated on $3$-way marginals with privacy budgets $\varepsilon \in \{ 0.1, 0.15, 0.2, 0.25, 0.5, 1 \}$ and $\delta = \frac{1}{n^2}$. The \textit{x-axis} uses a logarithmic scale. Results are averaged over $5$ runs, and error bars represent one standard error.
    }
    \label{fig:all_errors_mwem_no_opt}
\end{figure}

%% file: figures/all_errors/adult_loans.tex
\begin{figure}[!t]
    \centering
    \includegraphics[width=\linewidth]{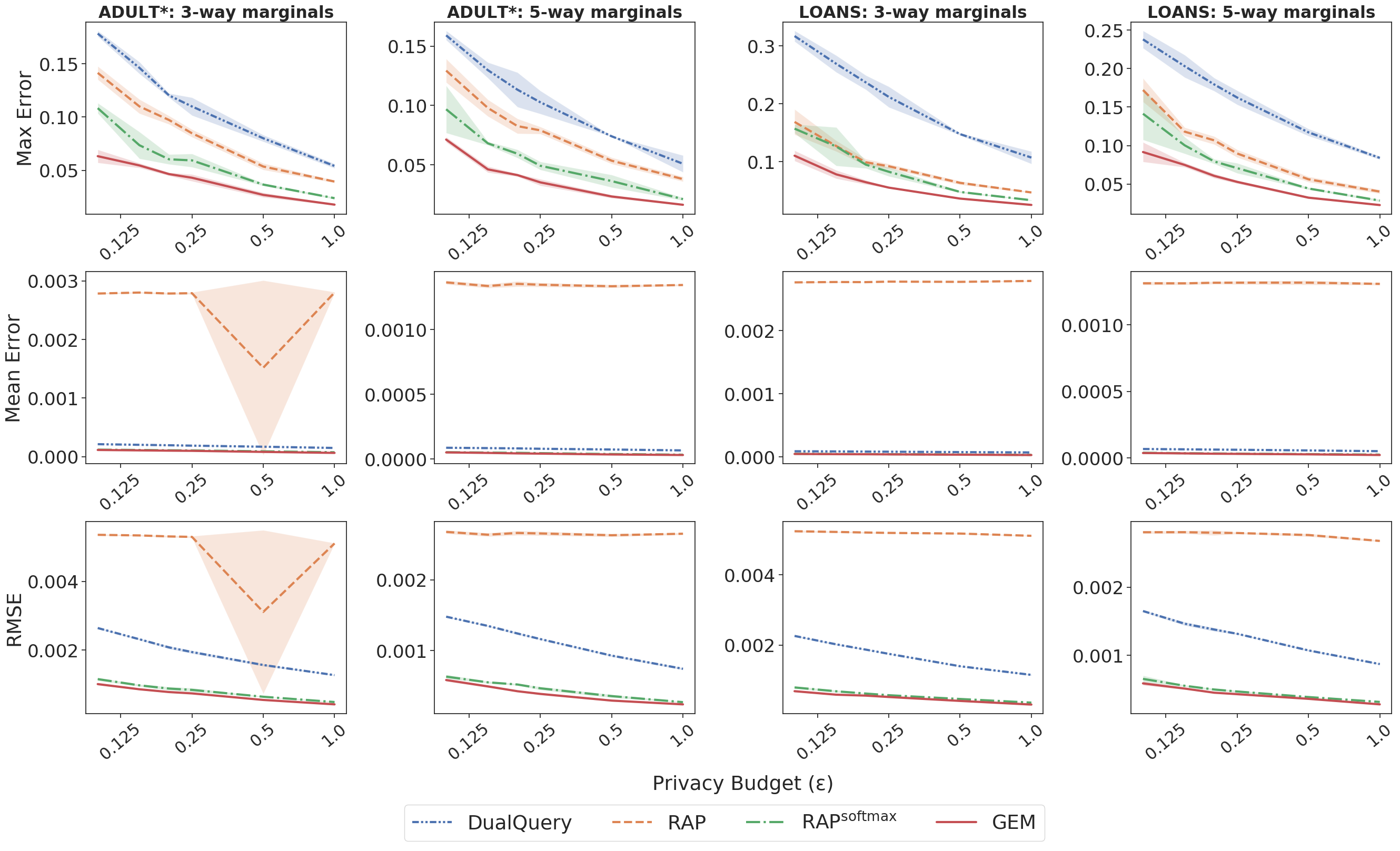}
    \caption{
    Max, mean, and root mean squared errors for $3$-way marginals with a workload size of $64$. Methods are evaluated on ADULT* and LOANS datasets using privacy budgets $\varepsilon \in \{ 0.1, 0.15, 0.2, 0.25, 0.5, 1 \}$ and $\delta = \frac{1}{n^2}$. The \textit{x-axis} uses a logarithmic scale. Results are averaged over $5$ runs, and error bars represent one standard error.
    }
    \label{fig:all_errors_adult_loans}
\end{figure}